\documentclass{article}

\usepackage[preprint]{neurips_2022}

\usepackage[utf8]{inputenc} 
\usepackage[T1]{fontenc}    
\usepackage{hyperref}       
\usepackage{url}            
\usepackage{booktabs}       
\usepackage{amsfonts}       
\usepackage{nicefrac}       
\usepackage{microtype}      
\usepackage{xcolor}         

\usepackage{bbm}
\usepackage{bm}
\usepackage{multirow}
\usepackage{booktabs}
\usepackage{graphicx}
\usepackage{caption}
\usepackage{subcaption}
\usepackage{textcomp}
\usepackage{enumitem}
\usepackage{pifont}
\usepackage{amsmath}
\usepackage[normalem]{ulem}
\usepackage{makecell}

\title{Unsupervised Text Style Transfer with Deep Generative Models}

\author{
Zhongtao Jiang$^{1,2}$, Yuanzhe Zhang$^{1,2}$, Yiming Ju$^{1,2}$, and Kang Liu$^{1,2}$\\
$^1$The Laboratory of Cognition and Decision Intelligence for Complex Systems,\\
Institute of Automation, Chinese Academy of Sciences\\
$^2$School of Artificial Intelligence, University of Chinese Academy of Sciences\\
\texttt{\{zhongtao.jiang, yzzhang, yiming.ju, kliu\}@nlpr.ia.ac.cn}\\
}

\begin{document}

\maketitle

\begin{abstract}
We present a general framework for unsupervised text style transfer with deep generative models. The framework models each sentence-label pair in the non-parallel corpus as partially observed from a complete quadruplet which additionally contains two latent codes representing the content and style, respectively. These codes are learned by exploiting dependencies inside the observed data. Then a sentence is transferred by manipulating them. Our framework is able to unify previous embedding and prototype methods as two special forms. It also provides a principled perspective to explain previously proposed techniques in the field such as aligned encoder and adversarial training. We further conduct experiments on three benchmarks. Both automatic and human evaluation results show that our methods achieve better or competitive results compared to several strong baselines\footnote{Code is available at \url{https://github.com/changmenseng/text_style_transfer}.}.
\end{abstract}

\section{Introduction}

Text generation is a key technology that facilitates many natural language processing (NLP) tasks, like language modeling \citep{radford-etal-2019-language, brown-etal-2020-language}, reply generation \citep{serban-etal-2016-building, serban-etal-2017-hierarchical}, etc. Recently, this field has made great strides, thanks to deep neural networks \citep{hochreiter-etal-1997-long, cho-etal-2014-learning, vaswani-etal-2017-attention}. Though current systems can generate fluent and plausible sentences, they are usually criticized for being uncontrollable \citep{fu-etal-2018-style}.

Towards controlled text generation, many efforts have been put in the task of text style transfer, which aims at controlling the style attributes of text while preserving its content \citep{jin-etal-2020-deep}, for example, flipping the sentiment polarity of a restaurant review. The main challenge is the lack of parallel corpus, preventing us from supervised learning. Thus, this paper mainly focuses on unsupervised text style transfer \citep{john-etal-2018-disentangled}. Existing approaches could be grouped into two main streams. The first one treats the task as an unsupervised machine translation problem \citep{lample-etal-2018-multiple, zhang-etal-2018-style, he-etal-2020-probabilistic}. It learns the transfer model directly by back-translation. For this stream \citet{he-etal-2020-probabilistic} has proposed an elegant probabilistic framework to explain different techniques. The second one tries to explicitly disentangle\footnote{Note that here the meaning of term ``disentanglement'' is different from that in unsupervised representation learning which typical exploits dimension-wise independence \citep{chen-etal-2018-isolating} in the code.} the content and style as two latent codes from a sentence and transfer a sentence by manipulating these codes. The code is either formed as a continuous embedding \citep{hu-etal-2017-toward,shen-etal-2017-style,fu-etal-2018-style, john-etal-2018-disentangled, wang-etal-2019-controllable, nangi-etal-2021-counterfactuals} or a text skeleton named prototype \citep{li-etal-2018-delete, sudhakar-etal-2019-transforming}. However, these two forms are usually considered completely different approaches\footnote{In fact, usually only the first form is named disentanglement method, while the second form is named prototype editing. In section \ref{sec:disentanglement} we show how prototype editing is categorized in disentanglement.}, and no works try to unify them to the best of our knowledge.

The core challenge of disentanglement-based methods is how to model these codes in the sentence since there are no annotations in the non-parallel corpora. This motivates us to utilize deep generative models \citep{kingma-etal-2013-auto, kingma-etal-2014-semi, rezende-etal-2014-stochastic} which take them as latent variables. Generative models are able to accommodate flexible dependency assumptions of variables in the data, where the latent variable could be learned implicitly by exploiting this structure. So far, generative models have been widely adopted in this field. However, existing approaches either fail to define a plausible generative structure (see Section \ref{sec:emb_form_background} and Appendix) or adopt heuristic objectives that lack theoretical support.

Therefore, we define a rational structure that treats the style label and content code as two generative factors of a sentence and derive objectives according to the principle of variational inference \citep{kingma-etal-2013-auto}. It turns out that we could unify the aforementioned two disentanglement-based approaches. Specifically, embedding and prototype forms could be regarded as the continuous and discrete cases in the proposed framework, respectively. Meanwhile, by diving into the variational objective, the framework theoretically explains previous techniques, such as aligned autoencoder \citep{shen-etal-2017-style} and adversarial loss \citep{romanov-etal-2018-adversarial, john-etal-2018-disentangled, bao-etal-2019-generating}. We conduct experiments on three datasets including \textsc{Yelp}, \textsc{Amazon} and \textsc{Captions} \citep{li-etal-2018-delete}. Both automatic and human evaluation results show that our methods achieve better or competitive results compared to several strong baselines in terms of style strength, content preservation, and language quality.

\section{Background \& Related Works}
\subsection{Unsupervised Text Style Transfer}\label{sec:transfer_background}

Parallel corpora are very scarce in the field, so most works study unsupervised text style transfer, where only style labeled corpus ${\cal D}=\{(x_1,y_1),(x_2,y_2),\cdots,(x_N,y_N)\}$ is available. Here, $x_i$ is a sentence and $y_i\in\{1,2,\cdots,Y\}$ is its style label. The goal is to learn a model $p(x'|x,y')$, which accepts a sentence $x$ and a target style $y'$ and generates the transferred sentence $x'$. The new sentence $x'$ is demanded to be in style $y'$ and preserve the content of original one $x$ at the same time. Here the content could be the theme or object, while the style could be sentiment (positive/negative) or writing style (romantic/humorous) in the sentence \citep{jin-etal-2020-deep}. $p(x'|x,y')$ is nontrivial to learn directly due to the lack of parallel corpus.

\subsection{Disentanglement}
\label{sec:disentanglement}
Formally, according to our observations, disentanglement methods generally de-marginalize the intractable $p(x'|x,y')$ into two models by introducing an additional variable $c$ that represents content:
\begin{equation}
p(x'|x,y')=\int p(c|x)p(x'|c,y')dc=\Bbb{E}_{p(c|x)}[p(x'|c,y')]
\label{eq:disentanglement_form}
\end{equation}
That is, they first obtain a content code $c$ from the encoder $p(c|x)$, then generate the objective sentence by a decoder $p(x'|c,y')$. The point is that $c$ must be informative to the content and agnostic to the style label $y$. Proposed approaches often form $c$ as either a continuous embedding or a prototype.

\subsubsection{Embedding Form}
\label{sec:emb_form_background}
Embedding form methods define the cotent code $c$ as a continuous embedding and are further divided into two categories according to how to manipulate it.

\paragraph{Attribute Code Control (ACC)} \citep{shen-etal-2017-style, fu-etal-2018-style, prabhumoye-etal-2018-style, yang-etal-2018-unsupervised, logeswaran-etal-2018-content} These methods directly apply Equation (\ref{eq:disentanglement_form}) that first refines a vector $c$ using a encoder $p(c|x)$. To make $c$ only carry information about content, adversarial learning is commonly applied to make $c$ be agnostic to $y$. Then they generate $x'$ with a decoder $p(x'|c,y')$. 

\paragraph{Representation Splitting (RS)} \citep{hu-etal-2017-toward, john-etal-2018-disentangled, romanov-etal-2018-adversarial} Besides $p(c|x)$, methods in this framework have another encoder $p(s|x)$ extracting a fine-gainer style code $s$ of $x$. The information in $x$ is then separated into the content code $c$ and style code $s$. The style is then modified by keeping $c$ only and altering $s$. To be concrete, they model the decoder $p(x'|c,y')$ as:
\begin{equation}
p(x'|c,y')={\Bbb E}_{p(s'|y')}[p(x'|s',c)],\quad p(s'|y')={\Bbb E}_{p_{\cal D}(x''|y')}[p(s'|x'')]
\end{equation}
where $p_{\cal D}(\cdot)$ is data empirical distribution. That is, they change the original style code $s$ to the average style code of target-style sentences. Then the transferred sentence is generated according to it and the original content code $c$. This framework is problematic in its assumed dependency structure, which treats $s$ and $c$ as two independent generative factors of $x$. This is irrational because stylistic expressions (represented by $s$) usually have a strong relatedness or dependence with the content pieces (represented by $c$), such as ``delicious'' for ``food'' and ``patient'' for ``service'' in sentiment transfer scenario. This shortcoming is further exacerbated such that many works arbitrarily assume the prior $p(s)$ as a uni-modal Gaussian \citep{john-etal-2018-disentangled, bao-etal-2019-generating, nangi-etal-2021-counterfactuals}, where it's factually more appropriate to be multi-modal. By contrast, as we will see, our proposed generative structure is more plausible and avoids setting this complex multi-modal prior. More details of RS framework are shown in Appendix.


\subsubsection{Prototype Form}\label{sec:prototype-editing}
Prototype form methods define $c$ as a prototype consisting of incomplete input sentence pieces. The starting point is that style is often marked by distinctive phrases in the sentence \citep{li-etal-2018-delete, madaan-etal-2020-politeness}. Therefore, simply deleting these markers would end up with a content-only prototype. The transferred sentence is therefore obtained by infilling the prototype with phrases of the target style. This pipeline is simple and effective. It clearly shows how the sentence is modified, which brings more interpretability. Previous works commonly obtain the content prototype heuristically, e.g., by thresholding importance scores, which come from frequency ratios of n-grams \citep{li-etal-2018-delete} or attention weights \citep{zhang-etal-2018-learning, sudhakar-etal-2019-transforming}. However, some importance scores such as attention are argued not explainable as expected \citep{jain-etal-2019-attention, serrano-etal-2019-attention, bastings-etal-2020-elephant}, and we need to set an annoying threshold. Moreover, previous approaches are often based on intuitions and don't realize the connection (Equation (\ref{eq:disentanglement_form})) with embedding form methods in Section \ref{sec:emb_form_background}.

\section{Framework}\label{sec:framework}
According to the formulation established in Equation (\ref{eq:disentanglement_form}), our goal is to learn an encoder $p(c|x)$ and a decoder $p(x|c,y)$ from a non-parallel corpus. Directly modeling these two distributions is infeasible since there are no annotations of the content code $c$. Instead, we treat $c$ as a latent variable and learn it by exploiting dependencies between the sentence $x$ and style label $y$. Concretely, we maximize the joint likelihood $p(x,y)$, which could be modeled in two ways:
\begin{equation}
    p(x,y)=p(x|y)p(y)=p(y|x)p(x)
\end{equation}
Since marginal distributions $p(x)$ and $p(y)$ can be learned directly from the data, we focus on learning remained two conditionals $p(x|y)$ and $p(y|x)$, where $c$ is implicitly introduced by these distributions.

To do this, we first follow RS approaches (Section \ref{sec:emb_form_background}) to introduce an additional latent variable $s$, which is viewed as a finer-grained code of style compared to the categorical label $y$. That is, $s$ could represent stylistic expressions in texts, such as sentiment phrases in sentiment transfer tasks. Then we define how a $(x,y,s,c)$-quadruplet is generated, i.e., the generative model. We assume the generative process to be: First $c$ and $y$ are sampled from prior distributions $p(c)$ and $p(y)$. Next $x$ is sampled from a decoder $p_\theta(x|c,y)$ which is built by a neural network with parameter $\theta$. At last $s$ is inferred by $p(s|x)$. This is depicted in Figure \ref{fig:generative_pgm}. Accordingly, the joint distribution is factorized as:
\begin{equation}
p_\theta(x,y,s,c)= p(y)p(c)p_\theta(x|c,y)p(s|x)
\end{equation}
We denote this generative model by ${\Bbb P}$. Essentially, we assume the style label $y$ and content code $c$ are two independent generative factors of a sentence $x$ \citet{kingma-etal-2014-semi}, which is plausible for text style transfer. Also, note that $p_\theta(x,y,c)$ is invariant to marginalization over $s$, which significantly simplifies the problem such that $s$ is out of consider in modeling $p(x|y)$, as we will see.

\begin{figure}
    \centering
    \begin{subfigure}[b]{0.22\linewidth}
        \centering
        \includegraphics[width=\linewidth]{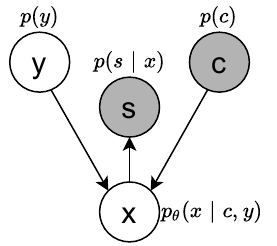}
        \caption{Generative Model ${\Bbb P}$.}
        \label{fig:generative_pgm}
    \end{subfigure}
    \begin{subfigure}[b]{0.22\linewidth}
        \centering
        \includegraphics[width=\linewidth]{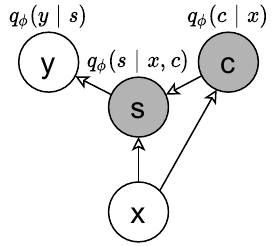}
        \caption{Inference Model ${\Bbb Q}$.}
        \label{fig:inference_pgm}
    \end{subfigure}
    \caption{Probabilistic graph of our framework, where shadowed nodes represent latent variables.}
    \label{fig:pgm}
\end{figure}

\subsection{ELBO Loss}
As presented, our first goal is to maximize the conditional likelihood $p_\theta(x|y)$ whose logarithm is bounded by the evidence lower bound (ELBO) \citep{kingma-etal-2013-auto}:

\begin{equation}
\begin{aligned}
\log p_\theta(x|y)\geq {\Bbb E}_{q_\phi(c|x,y)}\left[\log\frac{p(c)p_\theta(x|c,y)}{q_\phi(c|x,y)}\right]
={\Bbb E}_{q_\phi(c|x,y)}[\log p_\theta(x|c,y)]-\text{KL}(q_\phi(c|x,y)\Vert p(c))\\
\equiv -{\cal L}_{\theta,\phi}^\text{ELBO}(x,y)
\end{aligned}
\label{eq:ori_elbo_loss}
\end{equation}
where $q_\phi(c|x,y)$ is the variational approximation of the true posterior $p_\theta(c|x,y)$ with parameter $\phi$. We further assume that a sentence contains all the information about its content. Thus the content code $c$ is conditionally independent with the style $y$ given the sentence $x$ \citep{john-etal-2018-disentangled, romanov-etal-2018-adversarial, bao-etal-2019-generating}: $q_\phi(c|x,y)= q_\phi(c|x)$. Correspondingly, the ELBO loss is:
\begin{equation}
{\cal L}^{\text{ELBO}}_{\theta,\phi}(x,y)=-{\Bbb E}_{q_\phi(c|x)}[\log p_\theta(x|c,y)]+{\rm KL}(q_\phi(c|x)\Vert p(c))
\label{eq:elbo_loss}
\end{equation}
The first term ensures that the extracted content code $c$ along with the style label $y$ is capable to reconstruct the original sentence $x$, forcing $c$ to encode content information in $x$. This is often termed as a cycle reconstruction objective \citep{prabhumoye-etal-2018-style, logeswaran-etal-2018-content,luo-etal-2019-dual, jin-etal-2020-deep} and plays an important role in content preserving. The second term is the Kullback-Leibler (KL) divergence between the posterior $q_\phi(c|x)$ and the prior $p(c)$. It makes the extracted code entail the prior assumption for regularization.

\subsection{Classification Loss}
Our second goal is to maximize the conditional likelihood $p_\theta(y|x)$. Since exactly computing this probability is intractable, we introduce another tractable classifier $q_\phi(y|x)$ to approximate it. Although optimizing the surrogate $q_\phi(y|x)$ doesn't necessarily optimize the true distribution $p_\theta(y|x)$, this technique is widely-used in semi-supervised learning scenario \citep{kingma-etal-2014-semi, maaloe-etal-2016-auxiliary}, and we will show it helps learn the encoder $q_\phi(c|x)$. 

We treat the content code $c$ and the style code $s$ as two intermediate variables of the inference from the sentence $x$ to style label $y$. The classifier $q_\phi(y|x)$ is therefore acquired by marginalizing these latent variables in $q_\phi(y,s,c|x)$ out, which we assume:
\begin{equation}
    q_\phi(y,s,c|x)=q_\phi(c|x)q_\phi(s|x,c)q_\phi(y|s)
\end{equation}
We denote this model by inference model ${\Bbb Q}$, as presented in Figure \ref{fig:inference_pgm}\footnote{Note that we reuse the variational posterior $q_\phi(c|x)$ in the generative model ${\Bbb P}$ as the encoder in the inference model ${\Bbb Q}$, which connects two models.}. Contrast to previous work that apply mean-field assumptions $q_\phi(s,c|x)=q_\phi(c|x)q_\phi(s|x)$ \citep{hu-etal-2017-toward, john-etal-2018-disentangled, romanov-etal-2018-adversarial}, we use a more general form because term $q_\phi(s|x,c)$ entails an intuition that the style code $s$ could be inferred by ``subtracting'' content code $c$ from the sentence $x$. In this way $c$ and $s$ are connected and can be enhanced by exploiting each other. We will show that this assumption is the key to unify prototype form methods. Accordingly, the logarithm likelihood becomes:
\begin{equation}
\log q_\phi(y|x)=\log{\Bbb E}_{q_\phi(c|x)q_\phi(s|x,c)}[q_\phi(y|s)]\geq {\Bbb E}_{q_\phi(c|x)q_\phi(s|x,c)}[\log q_\phi(y|s)] \equiv -{\cal L}_{\theta,\phi}^{\rm CLS}(x,y)
\label{eq:classification_loss}
\end{equation}
We call this objective as the classification loss given that it is intrinsically learning a classifier. This loss makes the style code $s$ informative towards style label $y$ by pushing the content code $c$ out of the sentence $x$, which helps learn the encoder $q_\phi(c|x)$.

\subsection{Training}
Combining above objectives together, the total loss on the corpus ${\cal D}$ is defined as:
\begin{equation}
{\cal L}_{\theta,\phi}\equiv{\Bbb E}_{p_{\cal D}(x,y)}[{\cal L}^{\rm ELBO}_{\theta,\phi}(x,y)+\gamma{\cal L}^{\rm CLS}_{\theta,\phi}(x,y)]\simeq\frac{1}{N}\sum_{i=1}^N{\cal L}_{\theta,\phi}^{\rm ELBO}(x_i,y_i)+\gamma{\cal L}_{\theta,\phi}^{\rm CLS}(x_i,y_i)
\label{eq:total_loss}
\end{equation}
where $\gamma>0$ is a balancing hyperparameter.

\section{Connection to Previous Work}
\label{sec:connection}
Our framework could be connected with recent techniques in disentanglement methods for text style transfer. For example, the cycle reconstruction loss appears naturally in Equation (\ref{eq:elbo_loss}). Moreover, we show that aligning content code $c$ of across styles \citep{shen-etal-2017-style, shang-etal-2019-semi} could be derived from our framework in a principled way.

To see this, we decompose the ELBO loss as:
\begin{equation}
\begin{aligned}
{\cal L}_{\theta,\phi}^{\rm ELBO}&=\underbrace{-{\Bbb E}_{p_{\cal D}(x,y)q_\phi(c|x)}[\log p_\theta(x|c,y)]}_\text{\ding{172} Average reconstruction loss}\underbrace{+{\Bbb E}_{p_{\cal D}(y)}[I_{\tilde{q}_\phi(c,x|y)}(x;c)]}_\text{\ding{173} Average index-code MI along style}\\
&\underbrace{+\text{KL}(\tilde{q}_\phi(c)\Vert p(c))}_\text{\ding{174} Marginal KL to prior}\underbrace{+\text{JS}(\Vert_y\tilde{q}_\phi(c|y))}_\text{\ding{175} Aligning content code across styles}
\end{aligned}
\label{eq:elbo_decompose}
\end{equation}
where
\begin{equation}
\begin{aligned}
    \tilde{q}_\phi(c,x|y)=q_\phi(c|x)p_{\cal D}(x|y),\quad\tilde{q}_\phi(c|y)={\Bbb E}_{p_{\cal D}(x|y)}[q_\phi(c|x)],\quad\tilde{q}_\phi(c)={\Bbb E}_{p_{\cal D}(y)}[\tilde{q}_\phi(c|y)]
\end{aligned}
\end{equation}
Derivation is presented in Appendix. Here, $\tilde{q}_\phi(c|y)$ and $\tilde{q}_\phi(c)$ are aggregated posteriors \citep{makhzani-etal-2015-adversarial, tomczak-etal-2018-vae}. In this decomposition, term \ding{172} is the average reconstruction loss on the dataset. Term \ding{173} is called index-code mutual information (MI) \citep{hoffman-etal-2016-elbo}, where it encourages the code $c$ to cover distinct sentences $x$ as much as possible. Term \ding{174} is for regularization which tries to match the marginal aggregated posterior $\tilde{q}_\phi(c)$ to the prior $p(c)$. Term \ding{175} is the general Jensen-Shannon divergence of all aggregated posteriors conditioned on $y$ that tries to align distributions of content code $c$ in different styles. We consider this implicit alignment as the core technique in disentanglement approaches. This term is first intuitively introduced in \citet{shen-etal-2017-style}, but our framework arrives at it theoretically as a sub-objective of the ELBO loss. Directly computing this term is intractable. Instead, it's usually optimized via adversarial training \citep{goodfellow-etal-2014-generative, nowozin-etal-2016-f} that trains the encoder $q_\phi(c|x)$ to fool an adversary discriminator. While adversarial training is often explained to remove the style information in the code $c$ \citep{fu-etal-2018-style, yang-etal-2018-unsupervised, john-etal-2018-disentangled, jin-etal-2020-deep}, it essentially aligns distributions of $c$ across styles from our probabilistic view.

The proposed framework is twinned with \citet{he-etal-2020-probabilistic}, which our strengths and weaknesses are complementary. The advantage of our framework is that it could explicitly model content and style with careful treatment, which is natural to unify different forms of variables as we will see. However, our framework is not suitable for transduction tasks such that sentences of different domains are literally different such as word substitution decipherment \citep{shen-etal-2017-style, yang-etal-2018-unsupervised} and machine translation \citep{he-etal-2020-probabilistic} because in those cases using a single encoder for all domains is not proper anymore.

\section{Instantiation}\label{sec:instantiation}
In application, we need to design five components consisted in the framework, including a prior $p(c)$ and four parametric models: 1) decoder $p_\theta(x|c,y)$, 2) encoder $q_\phi(c|x)$, 3) auxiliary encoder $q_\phi(s|x,c)$, 4) classifier $q_\phi(y|s)$. In this section, we present two instantiations that respectively form the code $c$ as an embedding or a prototype. It shows that our framework could unify both forms.

\subsection{Embedding Form}
In this case, $c$ and $s$ are continuous variables: $c\in {\Bbb R}^{d_c}$, $s\in {\Bbb R}^{d_s}$. Components includes:

\begin{description}[style=unboxed,leftmargin=0.2cm]
\item[- \textnormal{\textsl{Prior $p(c)$}}]
The prior of $c$ is defined as an isotropic unit Gaussian distribution: $p(c)={\cal N}(0,I)$.
\item[- \textnormal{\textsl{Decoder $p_\theta(x|c,y)$}}] We build a long short term memory (LSTM) \citep{hochreiter-etal-1997-long} decoder for each style. We concatenate $c$ on input embeddings and hidden states of all time steps.

\item[- \textnormal{\textsl{Encoder $q_\phi(c|x)$}}]
The encoder is designed as a degenerate distribution $q_\phi(c|x)={\mathbbm 1}(c=c_\phi(x))$. $c_\phi(x)$ is a deterministic function built with a bidirectional LSTM (BiLSTM). We concatenate the last hidden states in both directions and predict the content code $c$ using a linear head.

\item[- \textnormal{\textsl{Auxiliary Encoder $q_\phi(s|x,c)$}}]
It is relatively hard to define a ``subtracting'' operation in the embedding space. Thus we leave this for future work and utilize previous settings that $c$ and $s$ are conditional independent of $x$, which gives $q_\phi(s|x,c)={\mathbbm 1}(s=s_\phi(x))$. $s_\phi(x)$ shares a same BiLSTM with $c_\phi(x)$ and uses another linear head for inference. So $c$ is also effected by $s$.

\item[- \textnormal{\textsl{Classifier $q_\phi(y|s)$}}]
The classifier is a linear head on $s$.

\end{description}

A LSTM decoder is a powerful auto-regressive model that could model the data distribution on its own \citep{radford-etal-2019-language}. To avoid posterior collapse \citep{bowman-etal-2015-generating}, we follow \citet{dieng-etal-2019-avoiding, zhao-etal-2017-infovae} to drop the index-code MI (term \ding{173}) in Equation (\ref{eq:elbo_decompose}) and obtain the final objective:
\begin{equation}
{\cal L}_{\theta,\phi}=-{\Bbb E}_{p_{\cal D}(x,y)}[\underbrace{\log p_\theta(x|c_\phi(x),y)}_{\text{\ding{172}}}+\underbrace{\log q_\phi(y|s_\phi(x))}_{{\cal L}_\phi^\text{CLS}}]+\underbrace{{\Bbb E}_{p_{\cal D}(y)}[\text{KL}(\tilde{q}_\phi(c|y)\Vert p(c))]}_{\text{\ding{174}}+\text{\ding{175}}}
\end{equation}
where the third KL term is unable to be computed analytically and we utilize adversarial training \citep{goodfellow-etal-2014-generative, makhzani-etal-2015-adversarial,  mohamed-etal-2016-learning, mescheder-etal-2017-adversarial} to approximate it. Details are shown in Appendix. 

\subsection{Prototype Form}
In this case, $c$ and $s$ are complementary text prototypes. The concept of the prototype is close to rationale, which is usually defined as input text pieces that are predictive towards some specific attributes \citep{lei-etal-2016-rationalizing, bastings-etal-2019-interpretable}. For example, for sentiment transfer task, the content and style prototype of a sentence ``I like this movie'' could be ``I \texttt{[MASK]} this movie'' and ``\texttt{[MASK]} like \texttt{[MASK]} \texttt{[MASK]}'', respectively. The prototype is determined by a binary mask $\bar{c}/\bar{s}\in\{0,1\}^{|x|}$ indicating presences of words in the prototype. Then a prototype is a sentence which masked positions take zero vectors as their word embeddings. We next introduce components in this case:

\begin{description}[style=unboxed,leftmargin=0.2cm]
\item[- \textnormal{\textsl{Prior $p(c)$}}] The prior of $c$ is defined as the distribution of its corresponding mask $\bar{c}$.
\begin{equation}
p(c)= p(\bar{c})
=p(|\bar{c}|)\prod_{t=1}^{|\bar{c}|}r_{|\bar{c}|}^{\bar{c}^{(t)}}(1-r_{|\bar{c}|})^{1-\bar{c}^{(t)}}
\end{equation}
where $p(|\bar{c}|)$ is the length probability. Here we follow \citet{chen-etal-2020-learning} that all the elements in the mask $\bar{c}$ are i.i.d Bernoulli variables with a length-dependent parameter, i.e., $\bar{c}^{(t)}\sim\text{Bern}(r_{|\bar{c}|})$.

\item[- \textnormal{\textsl{Decoder $p_\theta(x|c,y)$}}] Since $c$ is a sequence, for each style $y$ the decoder is parameterized using a sequence to sequence (seq2seq) RNN. Specifically, we utilize LSTMs with the attention mechanism \citep{bahdanau-etal-2014-neural}. We merge all the continuous masked words to one special \texttt{[MASK]} token for variable-length decoding \citep{lewis-etal-2019-bart, shen-etal-2020-blank}.

\item[- \textnormal{\textsl{Encoder $q_\phi(c|x)$}}]
Inferring $c$ is actually inferring $\bar{c}$ given that $c$ is a prototype. We further adopt mean-field approximation \citep{blei-etal-2017-variational} to posit the conditional independence:
\begin{equation}
q_\phi(c|x)=q_\phi(\bar{c}|x)=\prod_{t=1}^{|x|}{\mathbbm 1}(\bar{c}^{(t)}=\bar{c}_\phi^{(t)}(x))
\end{equation}
where $\bar{c}^{(t)}$ is the mask of $t$-th word in the sentence. $\bar{c}_\phi^{(t)}(x)$ is built with a BiLSTM followed by a linear head:
\begin{equation}
\begin{aligned}
h^{(t)}=\text{BiLSTM}_\phi^{(t)}(x^{(1)},\cdots,x^{(|x|)}), l^{(t)}=\sigma(Wh^{(t)}+b), \bar{c}^{(t)}={\mathbbm 1}(l^{(t)}>0.5)+l^{(t)}-\text{SG}(l^{(t)})
\end{aligned}
\end{equation}
Here, SG stands for a stop-gradient operator that is an identity at forward time and has zero derivatives. This straight through (ST) \citep{bengio-etal-2013-estimating} estimator brings us differential masks $\bar{c}$.

\item[- \textnormal{\textsl{Auxiliary Encoder $q_\phi(s|x,c)$}}]
To infer $s$, we also need to infer its mask $\bar{s}$. We consider $\bar{c}$ and $\bar{s}$ to be complementary following \citet{li-etal-2018-delete, sudhakar-etal-2019-transforming}. This gives:
\begin{equation}
q_\phi(s|x,c)= q(\bar{s}|{\bar{c}})={\mathbbm 1}(\bar{s}=1-\bar{c})
\end{equation}
As we can see, the auxiliary encoder has no parameters. $s$ is uniquely determined by subtracting $c$ from $x$: $s_\phi(x)=x\circ(1-\bar{c}_\phi(x))$. Clearly, this formulation could completely decouple $s$ and $c$ since they have no literal overlaps compared to learning the encoder $q_\phi(s|x)$ individually.

\item[- \textnormal{\textsl{Classifier $q_\phi(y|s)$}}] The classifier is a simple BiLSTM classifier. We concatenate last hidden representations in both directions and utilize a linear head to compose label probabilities.
\end{description}

In the specialization above, the total loss becomes:
\begin{equation}
{\cal L}_{\theta,\phi}=\underbrace{{\Bbb E}_{p_{\cal D}(x,y)}\big[-\log p_\theta(x|c_\phi(x),y)\big]}_{{\cal L}_{\theta,\phi}^{\rm INFILL}\text{: Learning to infill the content prototype}}+\quad\underbrace{{\Bbb E}_{p_{\cal D}(x,y)}\big[-\gamma\log q_\phi(y|s_\phi(x))+\frac{\alpha}{|x|}\Vert \bar{s}_\phi(x)\Vert_1\big]}_{{\cal L}_{\phi}^{\rm RAT}\text{: Learning the content prototype}}
\label{eq:mask_total_loss}
\end{equation}
where $\alpha=-|x|\log((1-r_{|x|})/r_{|x|})$ is a hyperparmeter. Derivation details are shown in Appendix. When we set $\alpha>0$, ${\cal L}_{\phi}^\text{RAT}$ is almost identical to the definition of the rationale \citep{lei-etal-2016-rationalizing, bastings-etal-2019-interpretable, de-etal-2020-decisions} except for the contiguity loss, where the first and second terms refer to fidelity and compactness loss respectively \citep{jiang-etal-2021-alignment}, two of the requirements of faithful and readable interpretations \citep{molnar-etal-2020-interpretable}. The training process is conducted in two steps. First we optimize ${\cal L}_{\phi}^{\rm RAT}$ to train a rationale system $\bar{c}_{\phi^\star}(x)$, and obtain the content prototype $c$ for each instance in ${\cal D}$. Then we train text infilling decoders $p_\theta(x|c,y)$ by optimizing ${\cal L}_{\theta,\phi^\star}^{\rm INFILL}$. This two-step pipeline exactly follows the workflow of prototype editing approaches reviewed in section \ref{sec:prototype-editing}. Therefore, the prototype form is unified in our framework. Compared to prior approaches, our method learns the prototype using interpretability principles instead of intuition.

\section{Experiments}

\subsection{Settings}
\label{sec:settings}
\paragraph{Datasets} We adopt three datasets for our experiments: \textsc{Yelp}, \textsc{Amazon} and \textsc{Captions} \citep{li-etal-2018-delete}. The former two datasets require a system to alter the sentiment of a sentence from positive/negative to negative/positive. \textsc{Captions} \citep{gan-etal-2017-stylenet} aims at changing the writing style of a sentence from factual to romantic/humorous. \citet{li-etal-2018-delete} creates a testing set with annotated style transferred human references for each dataset and we adopt their version.

\paragraph{Baselines} We compare embedding form methods and prototype form methods separately. We select embedding form baselines including method based on RS framework (see Appendix for details), ACC methods \textsc{CrossAlign} \citep{shen-etal-2017-style} and \textsc{MultiDecoder} \citep{fu-etal-2018-style}. For prototype form, we compare our method with the typical delete-and-generate architecture \citep{li-etal-2018-delete} that first deletes stylistic phrases in the sentence to obtain a prototype and then re-infills it. Baseline methods delete stylistic phrases using thresholding importance scores that come from different attribution explanation methods including phrase relative \textsc{Frequency} \citep{li-etal-2018-delete}, \textsc{Attention}, \textsc{Gradient} \citep{simonyan-etal-2013-deep}, \textsc{IntegratGrad} \citep{sundararajan-etal-2017-axiomatic} and LIME \citep{ribeiro-etal-2016-should}. For \textsc{Frequency}, we adopt thresholds of the original paper. For other attribution methods, importance scores are obtained by a classifier based on BiLSTM with attention pooling\footnote{The classifier achieves 98.22\%, 85.30\% and 78.90\% on \textsc{Yelp}, \textsc{Amazon} and \textsc{Captions} validation sets.}. We set the threshold as the average score on the sentence for other methods following \citet{xu-etal-2018-unpaired}, which leads to strong baselines. We don't remove the classification loss for ablation because our prototype form method cannot be derived without it. Towards fair comparison, all methods use the same architecture as much as possible. Detailed hyperparameter settings are shown in Appendix.

\paragraph{Evaluation} We evaluate a transfer system in three criteria: transfer strength, content preservation, and language quality following \citet{jin-etal-2020-deep}. We conduct both automatic evaluation and human evaluation. For automatic evaluation, transfer strength is mesured by the classification accuracy (ACC) of transferred sentences according to a pre-trained classifier. We tune a BERT-base \citep{devlin-etal-2019-bert} classifier on each dataset for this purpose\footnote{The classifier achieves 98.32\%, 88.15\% and 82.50\% on \textsc{Yelp}, \textsc{Amazon} and \textsc{Captions} validation sets.}. Content preservation is evaluated by BLEU$_s$ and BLEU$_r$. The former is the BLEU score \citep{papineni-etal-2002-bleu} between transferred and original sentences, while the latter is the BLEU score between transferred and renference sentences\footnote{BLEU$_r$ could also reflect the transfer strength, thus this is the most persuasive automatic metric.}. Language quality is evaluated using language perplexity (PPL). We train a 3-gram language model with Kneser-Ney smoothing using SRILM toolkit \citep{stolcke-etal-2002-srilm} on the training set of each dataset to do this. We also report the geometric mean (GM) of ACC, BLEU$_s$, BLEU$_r$ and $1/\log(\text{PPL})$ as the overall score. Each compared method adopts the checkpoint that maximizes the GM score on the validation set. For human evaluation, we recruit two raters with linguistic background to score the generated text from 1 to 5 in terms of the aforementioned criteria\footnote{For each system, we sample 100 instances in the testing set. The raters are agnostic to which
model the generated text comes from. The correlation coefficients of two raters are 0.81, 0.77 and 0.45 for transfer strength, content preservation, and language quality, respectively.}. The final human score (HS) of a sentence for each aspect is the average of their scores. We follow \citet{li-etal-2018-delete, luo-etal-2019-dual} to consider a successful transfer if all aspects are rated equal or above 4 and report the successful ratio (SUC).

\useunder{\uline}{\ul}{}
\begin{table}[]
\centering
\tiny
\caption{Automatic and human evaluation results. ``Y'', ``A'' and ``C'' stand for \textsc{Yelp}, \textsc{Amazon}, and \textsc{Captions} dataset, respectively. For each column, top 2 results are bold and underlined, respectively.\\}
\label{tab:eval_results}
\begin{tabular}{llccccccccc}
\toprule
\multirow{2}{*}{Task} & \multirow{2}{*}{Method} & \multicolumn{2}{c}{Transfer Strength} & \multicolumn{3}{c}{Content Preservation}        & \multicolumn{2}{c}{Language}   & \multicolumn{2}{c}{Summary}     \\ \cmidrule(r){3-4} \cmidrule(r){5-7} \cmidrule(r){8-9} \cmidrule(r){10-11}
                      &                         & ACC$\uparrow$                & HS$\uparrow$                & BLEU$_s\uparrow$        & BLEU$_r\uparrow$        & HS$\uparrow$             & PPL$\downarrow$            & HS$\uparrow$             & GM$\uparrow$             & SUC$\uparrow$            \\ \midrule
\multirow{12}{*}{Y}   & RS                      & 69.40              & 3.84      & 03.71          & 02.37          & 1.98          & \textbf{08.72} & \textbf{4.44} & 04.92          & 03.00          \\
                      & \textsc{CrossAlign}              & {\ul 73.40}        & 3.56             & 21.05          & 09.40          & {\ul 3.45}    & {\ul 38.98}    & 4.15          & 07.94          & {\ul 28.00}    \\
                      & \textsc{MultiDec}                & \textbf{73.90}     & 3.48             & 24.73          & 10.71          & 3.27          & 44.63          & 3.96          & 08.47          & 13.00          \\
                      & $\text{\ding{172}}+{\cal L}_\phi^\text{CLS}$ & 59.00 & 3.05 & \textbf{29.49} & \textbf{12.00} & 3.38 & 47.45 & 3.75 & 08.58 & 15.00 \\
                      & $\text{\ding{172}}+\text{\ding{174}}+{\cal L}_\phi^\text{CLS}$ & 68.70 & 3.55 & {\ul 26.36} & 11.28 & {\ul 3.43} & 40.13 & 3.63 & {\ul 08.63} & 20.00 \\
                      & $\text{\ding{172}}+\text{\ding{174}}+\text{\ding{175}}$ & 72.10              & {\ul 3.76}             & 25.47    & 11.26    & \textbf{3.49} & 42.13          & 4.21          & 08.62    & 23.00          \\
                      & $\text{\ding{172}}+\text{\ding{174}}+\text{\ding{175}}+{\cal L}_\phi^\text{CLS}$ (Ours)    & 72.70              & \textbf{3.88}    & 25.90 & {\ul 11.47} & 3.31          & 45.36          & {\ul 4.24}    & \textbf{08.67} & \textbf{30.00} \\ \cmidrule{2-11} 
                      & Frequency               & 78.90              & {\ul 4.16}       & {\ul 54.84}    & 24.47          & 4.01          & {\ul 46.72}    & {\ul 4.33}    & 12.88          & 40.00          \\
                      & \textsc{Attention}               & 84.10              & 3.56             & 54.37          & {\ul 26.06}    & {\ul 4.41}    & 51.68          & 4.17          & {\ul 13.18}    & {\ul 45.00}    \\
                      & \textsc{Gradient}                & 92.50              & 4.10             & 33.57          & 17.66          & 3.90          & \textbf{43.92} & 4.26          & 10.97          & 48.00          \\
                      & \textsc{IntegratGrad}            & {\ul 96.40}        & 3.99             & 31.08          & 17.01          & 3.77          & 46.05          & 4.03          & 10.74          & 33.00          \\
                      & LIME                    & \textbf{98.00}     & \textbf{4.31}    & 27.67          & 15.28          & 3.11          & 52.90          & 3.83          & 10.11          & 23.00          \\
                      & Ours (Prototype Form)        & 83.40              & 3.70             & \textbf{65.06} & \textbf{27.69} & \textbf{4.71} & 56.42          & \textbf{4.34} & \textbf{13.89} & \textbf{55.00} \\ \cmidrule{2-11} 
                      & Reference               & 78.60              & 4.78             & 32.92          & 100.0          & 4.78          & 108.6          & 4.85          & 15.33          & 80.00          \\ \midrule
\multirow{12}{*}{A}   & RS                      & 58.80              & \textbf{3.45}    & 04.29          & 02.34          & 2.15          & \textbf{13.47} & {\ul 4.22}    & 04.59          & 08.00          \\
                      & \textsc{CrossAlign}              & 66.20              & 3.01             & 21.97          & 11.60          & 2.59          & 37.66          & 4.06          & 08.26          & {\ul 13.00}    \\
                      & \textsc{MultiDec}                & 57.00              & 2.55             & \textbf{29.27} & \textbf{14.71} & 2.99   & 57.92          & 4.14          & \textbf{08.82} & 05.00          \\
                      & $\text{\ding{172}}+{\cal L}_\phi^\text{CLS}$ & 54.50 & 2.50 & {\ul 28.53} & {\ul 14.19} & {\ul 3.08} & 55.87 & 3.83 & {\ul 08.61} & 13.00 \\
                      & $\text{\ding{172}}+\text{\ding{174}}+{\cal L}_\phi^\text{CLS}$ & 64.50 & 2.88 & 22.76 & 11.71 & \textbf{3.08} & 38.54 & 3.70 & 08.28 & 12.00 \\
                      & $\text{\ding{172}}+\text{\ding{174}}+\text{\ding{175}}$   & {\ul 67.80}        & {\ul 3.24}       & 20.43          & 10.43          & 2.95          & {\ul 32.35}    & 4.18          & 08.03          & 10.00          \\
                      & $\text{\ding{172}}+\text{\ding{174}}+\text{\ding{175}}+{\cal L}_\phi^\text{CLS}$ (Ours)       & \textbf{67.80}     & 3.14             & 22.70    & 11.62    & 3.06 & 40.48          & \textbf{4.28} & 08.34    & \textbf{14.00} \\ \cmidrule{2-11} 
                      & \textsc{Frequency}               & 63.70              & 2.96             & \textbf{63.36} & \textbf{33.17} & 3.41          & 48.66          & \textbf{4.47} & \textbf{13.62} & \textbf{20.00} \\
                      & \textsc{Attention}               & 33.90              & 1.94             & {\ul 62.33}    & {\ul 32.97}    & \textbf{4.15} & 46.49          & 4.39          & {\ul 11.61}    & 13.00          \\
                      & \textsc{Gradient}                & 60.30              & {\ul 3.00}       & 29.53          & 16.82          & {\ul 3.54}    & {\ul 34.66}    & 4.05          & 09.59          & {\ul 18.00}    \\
                      & \textsc{IntegratGrad}            & {\ul 67.40}        & 2.66             & 28.15          & 16.00          & 3.12          & 36.23          & 4.12          & 09.59          & 13.00          \\
                      & LIME                    & \textbf{71.50}     & 2.67             & 26.13          & 14.09          & 2.86          & 44.59          & 4.24          & 09.12          & 10.00          \\
                      & Ours (Prototype Form)        & 60.10              & \textbf{3.01}    & 35.22          & 19.19          & 3.21          & \textbf{31.62} & {\ul 4.40}    & 10.41          & 15.00          \\ \cmidrule{2-11} 
                      & Reference               & 52.70              & 4.35             & 51.08          & 100.0          & 4.70          & 140.6          & 4.75          & 15.27          & 68.00          \\ \midrule
\multirow{12}{*}{C}   & RS                      & 60.00              & 3.50             & 10.57          & 04.60          & 2.12          & \textbf{13.39} & \textbf{4.22} & 06.25          & 05.00          \\
                      & \textsc{CrossAlign}              & 66.33              & 3.44             & \textbf{37.81} & \textbf{09.27} & \textbf{3.38} & 22.19          & 3.89          & \textbf{09.31} & 10.00          \\
                      & \textsc{MultiDec}                & 68.50              & 3.30             & {\ul 33.85}    & {\ul 08.78}    & 3.19    & 24.66          & 3.90          & 08.93   & 10.00          \\
                      & $\text{\ding{172}}+{\cal L}_\phi^\text{CLS}$ & 68.83 & 3.33 & 34.41 & 08.85 & {\ul 3.28} & 24.35 & 4.08 & {\ul 09.00} & 13.00 \\
                      & $\text{\ding{172}}+\text{\ding{174}}+{\cal L}_\phi^\text{CLS}$ & {\ul 84.67} & 3.63 & 24.77 & 07.74 & 2.84 & 18.47 & 4.12 & 08.64 & 15.00 \\
                      & $\text{\ding{172}}+\text{\ding{174}}+\text{\ding{175}}$       & 81.15        & {\ul 3.67}       & 26.70          & 07.79          & 2.90          & {\ul 16.81}    & {\ul 3.92}    & 08.80          & {\ul 15.00}    \\
                      & $\text{\ding{172}}+\text{\ding{174}}+\text{\ding{175}}+{\cal L}_\phi^\text{CLS}$ (Ours)     & \textbf{86.17}     & \textbf{3.83}    & 24.70          & 08.51          & 3.01          & 19.84          & 3.81          & 08.82          & \textbf{15.00} \\ \cmidrule{2-11} 
                      & \textsc{Frequency}               & 76.50              & 3.61             & \textbf{49.71} & {\ul 14.03}    & {\ul 3.60}    & 19.27          & 4.09          & {\ul 11.59}    & 10.00          \\
                      & \textsc{Attention}               & \textbf{88.67}     & 4.10             & 34.31          & 12.00          & 3.25          & {\ul 14.33}    & {\ul 4.15}    & 10.82          & {\ul 23.00}    \\
                      & \textsc{Gradient}                & 87.17              & 3.77             & 32.57          & 10.97          & 3.24          & 14.60          & 3.85          & 10.38          & 18.00          \\
                      & \textsc{IntegratGrad}            & 80.83              & 3.73             & 34.73          & 11.41          & 3.18          & 14.87          & 3.94          & 10.44          & 15.00          \\
                      & LIME                    & 88.00              & {\ul 4.17}       & 31.96          & 11.26          & 3.19          & \textbf{14.01} & 4.00          & 10.46          & 15.00          \\
                      & Ours (Prototype Form)        & {\ul 88.17}        & \textbf{4.19}    & {\ul 45.32}    & \textbf{15.29} & \textbf{3.96} & 23.52          & \textbf{4.17} & \textbf{11.79} & \textbf{40.00} \\ \cmidrule{2-11} 
                      & Reference               & 82.67              & 4.67             & 19.44          & 100.0          & 4.28          & 53.58          & 4.75          & 14.17          & 68.00          \\ \bottomrule
\end{tabular}
\end{table}


\subsection{Results}
Table \ref{tab:eval_results} shows results of different methods on three datasets. We obtain the following findings.

By comparing embedding form methods, firstly, our method outperforms or matches previous approaches. Specifically, for automatic evaluation, it obtains the best and second-best GM score on \textsc{Yelp} and \textsc{Amazon} respectively. For human evaluation, our method performs the best successful transfer ratio across datasets. Notably, our method surpasses the RS-based method with big margins, which demonstrates the superiority of our probabilistic assumptions. Secondly, our approach achieves a better balance between transfer strength and content preservation. For example, although other baselines look better on \textsc{Captions} in terms of content preservation, they pay this at a huge cost of transfer strength which our approach exceeds with about 20 points of accuracy. Human evaluation results also illustrate this, while our method has a better successful ratio. We also find that our method truly disentangles content and style, while style vectors are discriminative for the style and content vectors are not, as shown in Appendix.

By comparing prototype form methods, our method achieves the best or competitive results on most metrics for \textsc{Yelp} and \textsc{Captions}. Especially, it significantly surpasses other methods in terms of successful ratio. However, our method is of inferiority on \textsc{Amazon}. Our speculation is that style attributes are generally expressed obscurely and variously in this dataset rather than explicitly or directly by some simple expressions like ``good'' in \textsc{Yelp}, e.g., a positive review ``for the average kitchen i suspect this would last a lifetime'' in \textsc{Amazon}. For these sentences, our method tends to over-mask the sentence to get a content prototype. This explains why its transfer strength is good while content preservation is relative poor according to human scores. On the contrary, for a superior method such as \textsc{Frequency}, the masking operation is more cautious. In fact, for \textsc{Frequency}, we find that 33.2\% of content prototypes are identical to their original sentences on the testing set of \textsc{Amazon}. That is, the system doesn't delete any stylistic phrases in these sentences such that their transferred version is just a copy of them. Thus, it's not surprising that such outputs achieve the best content preservation.

Last, by comparison across forms, prototype form methods are significantly better than embedding form methods in general, which is consistent with previous work \citep{li-etal-2018-delete}. It shows that many sentences can be transferred by only changing some stylistic phrases for these tasks. We present a few outputs of different systems in Appendix.

\subsection{Ablation Study}
Since there are four terms in our embedding form method ($\text{\ding{172}}+\text{\ding{174}}+\text{\ding{175}}+{\cal L}_\phi^\text{CLS}$), we are curious about how these terms help in our task. We first remove ${\cal L}_\phi^\text{CLS}$ and find that almost all metrics get worse after this, verifying its effectiveness and necessity. Then we inspect \ding{174} and \ding{175}, which are regularization terms of the content code $c$. We find that on almost all datasets, the transfer strength is keep falling and the content preservation conversely keeps improving when we remove \ding{174} and \ding{175} in turn. This phenomenon is reasonable, since \ding{175} punishes differences among $\tilde{q}_\phi(c|y), y\in \{1,2,\cdots,Y\}$, which removing it makes the content code $c$ to be differently distributed across styles. As a result, the decoder $p_\theta(x|c,y)$ is feed with unseen vectors. This out of distribution (OOD) issue prevents the decoder from generating sentence with correct style $y$. Reserving \ding{174} alleviates this problem in a bit, given it requires the content code $c$ to be embedded in an uni-modal unit Gaussian manifold. Therefore, content codes $c$ from different styles at least have some overlaps. From another point, removing \ding{174} and \ding{175} makes the model similar to an autoencoder which reconstructs the original sentence very well, leading to a better content preservation but poor transfer strength. Overall it's worth adding these regularization terms because it leads better successful rate.

\section{Conclusion}
We propose a general framework based on deep generative models for text style transfer. The framework is able to unify previous embedding and prototype methods as two special forms. It also provides a new perspective to explain proposed techniques such as aligning content code across styles and adversarial training. We demonstrate the effectiveness of our approach via experiments and human evaluation.

\bibliographystyle{neurips_2022}
\bibliography{ref}

\begin{thebibliography}{69}
\providecommand{\natexlab}[1]{#1}
\providecommand{\url}[1]{\texttt{#1}}
\expandafter\ifx\csname urlstyle\endcsname\relax
  \providecommand{\doi}[1]{doi: #1}\else
  \providecommand{\doi}{doi: \begingroup \urlstyle{rm}\Url}\fi

\bibitem[Bahdanau et~al.(2014)Bahdanau, Cho, and
  Bengio]{bahdanau-etal-2014-neural}
Bahdanau, D., Cho, K., and Bengio, Y.
\newblock Neural machine translation by jointly learning to align and
  translate.
\newblock \emph{arXiv preprint arXiv:1409.0473}, 2014.

\bibitem[Bao et~al.(2019)Bao, Zhou, Huang, Li, Mou, Vechtomova, Dai, and
  Chen]{bao-etal-2019-generating}
Bao, Y., Zhou, H., Huang, S., Li, L., Mou, L., Vechtomova, O., Dai, X., and
  Chen, J.
\newblock Generating sentences from disentangled syntactic and semantic spaces.
\newblock \emph{arXiv preprint arXiv:1907.05789}, 2019.

\bibitem[Bastings \& Filippova(2020)Bastings and
  Filippova]{bastings-etal-2020-elephant}
Bastings, J. and Filippova, K.
\newblock The elephant in the interpretability room: Why use attention as
  explanation when we have saliency methods?
\newblock \emph{arXiv preprint arXiv:2010.05607}, 2020.

\bibitem[Bastings et~al.(2019)Bastings, Aziz, and
  Titov]{bastings-etal-2019-interpretable}
Bastings, J., Aziz, W., and Titov, I.
\newblock Interpretable neural predictions with differentiable binary
  variables.
\newblock \emph{arXiv preprint arXiv:1905.08160}, 2019.

\bibitem[Bengio et~al.(2013)Bengio, L{\'e}onard, and
  Courville]{bengio-etal-2013-estimating}
Bengio, Y., L{\'e}onard, N., and Courville, A.
\newblock Estimating or propagating gradients through stochastic neurons for
  conditional computation.
\newblock \emph{arXiv preprint arXiv:1308.3432}, 2013.

\bibitem[Blei et~al.(2017)Blei, Kucukelbir, and
  McAuliffe]{blei-etal-2017-variational}
Blei, D.~M., Kucukelbir, A., and McAuliffe, J.~D.
\newblock Variational inference: A review for statisticians.
\newblock \emph{Journal of the American statistical Association}, 112\penalty0
  (518):\penalty0 859--877, 2017.

\bibitem[Bowman et~al.(2015)Bowman, Vilnis, Vinyals, Dai, Jozefowicz, and
  Bengio]{bowman-etal-2015-generating}
Bowman, S.~R., Vilnis, L., Vinyals, O., Dai, A.~M., Jozefowicz, R., and Bengio,
  S.
\newblock Generating sentences from a continuous space.
\newblock \emph{arXiv preprint arXiv:1511.06349}, 2015.

\bibitem[Brown et~al.(2020)Brown, Mann, Ryder, Subbiah, Kaplan, Dhariwal,
  Neelakantan, Shyam, Sastry, Askell, et~al.]{brown-etal-2020-language}
Brown, T.~B., Mann, B., Ryder, N., Subbiah, M., Kaplan, J., Dhariwal, P.,
  Neelakantan, A., Shyam, P., Sastry, G., Askell, A., et~al.
\newblock Language models are few-shot learners.
\newblock \emph{arXiv preprint arXiv:2005.14165}, 2020.

\bibitem[Chen \& Ji(2020)Chen and Ji]{chen-etal-2020-learning}
Chen, H. and Ji, Y.
\newblock Learning variational word masks to improve the interpretability of
  neural text classifiers.
\newblock \emph{arXiv preprint arXiv:2010.00667}, 2020.

\bibitem[Chen et~al.(2018)Chen, Li, Grosse, and
  Duvenaud]{chen-etal-2018-isolating}
Chen, R.~T., Li, X., Grosse, R.~B., and Duvenaud, D.~K.
\newblock Isolating sources of disentanglement in variational autoencoders.
\newblock \emph{Advances in neural information processing systems}, 31, 2018.

\bibitem[Cho et~al.(2014)Cho, Van~Merri{\"e}nboer, Gulcehre, Bahdanau,
  Bougares, Schwenk, and Bengio]{cho-etal-2014-learning}
Cho, K., Van~Merri{\"e}nboer, B., Gulcehre, C., Bahdanau, D., Bougares, F.,
  Schwenk, H., and Bengio, Y.
\newblock Learning phrase representations using rnn encoder-decoder for
  statistical machine translation.
\newblock \emph{arXiv preprint arXiv:1406.1078}, 2014.

\bibitem[De~Cao et~al.(2020)De~Cao, Schlichtkrull, Aziz, and
  Titov]{de-etal-2020-decisions}
De~Cao, N., Schlichtkrull, M., Aziz, W., and Titov, I.
\newblock How do decisions emerge across layers in neural models?
  interpretation with differentiable masking.
\newblock \emph{arXiv preprint arXiv:2004.14992}, 2020.

\bibitem[Devlin et~al.(2019)Devlin, Chang, Lee, and
  Toutanova]{devlin-etal-2019-bert}
Devlin, J., Chang, M.-W., Lee, K., and Toutanova, K.
\newblock {BERT}: Pre-training of deep bidirectional transformers for language
  understanding.
\newblock In \emph{Proceedings of the 2019 Conference of the North {A}merican
  Chapter of the Association for Computational Linguistics: Human Language
  Technologies, Volume 1 (Long and Short Papers)}, pp.\  4171--4186,
  Minneapolis, Minnesota, June 2019. Association for Computational Linguistics.
\newblock \doi{10.18653/v1/N19-1423}.
\newblock URL \url{https://aclanthology.org/N19-1423}.

\bibitem[Dieng et~al.(2019)Dieng, Kim, Rush, and
  Blei]{dieng-etal-2019-avoiding}
Dieng, A.~B., Kim, Y., Rush, A.~M., and Blei, D.~M.
\newblock Avoiding latent variable collapse with generative skip models.
\newblock In \emph{The 22nd International Conference on Artificial Intelligence
  and Statistics}, pp.\  2397--2405. PMLR, 2019.

\bibitem[Fu et~al.(2018)Fu, Tan, Peng, Zhao, and Yan]{fu-etal-2018-style}
Fu, Z., Tan, X., Peng, N., Zhao, D., and Yan, R.
\newblock Style transfer in text: Exploration and evaluation.
\newblock In \emph{Proceedings of the AAAI Conference on Artificial
  Intelligence}, volume~32, 2018.

\bibitem[Gan et~al.(2017)Gan, Gan, He, Gao, and Deng]{gan-etal-2017-stylenet}
Gan, C., Gan, Z., He, X., Gao, J., and Deng, L.
\newblock Stylenet: Generating attractive visual captions with styles.
\newblock In \emph{Proceedings of the IEEE Conference on Computer Vision and
  Pattern Recognition}, pp.\  3137--3146, 2017.

\bibitem[Goodfellow et~al.(2014)Goodfellow, Pouget-Abadie, Mirza, Xu,
  Warde-Farley, Ozair, Courville, and Bengio]{goodfellow-etal-2014-generative}
Goodfellow, I., Pouget-Abadie, J., Mirza, M., Xu, B., Warde-Farley, D., Ozair,
  S., Courville, A., and Bengio, Y.
\newblock Generative adversarial nets.
\newblock \emph{Advances in neural information processing systems}, 27, 2014.

\bibitem[He et~al.(2020)He, Wang, Neubig, and
  Berg-Kirkpatrick]{he-etal-2020-probabilistic}
He, J., Wang, X., Neubig, G., and Berg-Kirkpatrick, T.
\newblock A probabilistic formulation of unsupervised text style transfer.
\newblock \emph{arXiv preprint arXiv:2002.03912}, 2020.

\bibitem[Hochreiter \& Schmidhuber(1997)Hochreiter and
  Schmidhuber]{hochreiter-etal-1997-long}
Hochreiter, S. and Schmidhuber, J.
\newblock Long short-term memory.
\newblock \emph{Neural computation}, 9\penalty0 (8):\penalty0 1735--1780, 1997.

\bibitem[Hoffman \& Johnson(2016)Hoffman and Johnson]{hoffman-etal-2016-elbo}
Hoffman, M.~D. and Johnson, M.~J.
\newblock Elbo surgery: yet another way to carve up the variational evidence
  lower bound.
\newblock In \emph{Workshop in Advances in Approximate Bayesian Inference,
  NIPS}, volume~1, 2016.

\bibitem[Hu et~al.(2017)Hu, Yang, Liang, Salakhutdinov, and
  Xing]{hu-etal-2017-toward}
Hu, Z., Yang, Z., Liang, X., Salakhutdinov, R., and Xing, E.~P.
\newblock Toward controlled generation of text.
\newblock In \emph{International Conference on Machine Learning}, pp.\
  1587--1596. PMLR, 2017.

\bibitem[Jain \& Wallace(2019)Jain and Wallace]{jain-etal-2019-attention}
Jain, S. and Wallace, B.~C.
\newblock Attention is not explanation.
\newblock \emph{arXiv preprint arXiv:1902.10186}, 2019.

\bibitem[Jiang et~al.(2021)Jiang, Zhang, Yang, Zhao, and
  Liu]{jiang-etal-2021-alignment}
Jiang, Z., Zhang, Y., Yang, Z., Zhao, J., and Liu, K.
\newblock Alignment rationale for natural language inference.
\newblock In \emph{Proceedings of the 59th Annual Meeting of the Association
  for Computational Linguistics and the 11th International Joint Conference on
  Natural Language Processing (Volume 1: Long Papers)}, pp.\  5372--5387, 2021.

\bibitem[Jin et~al.(2020)Jin, Jin, Hu, Vechtomova, and
  Mihalcea]{jin-etal-2020-deep}
Jin, D., Jin, Z., Hu, Z., Vechtomova, O., and Mihalcea, R.
\newblock Deep learning for text style transfer: A survey.
\newblock \emph{arXiv preprint arXiv:2011.00416}, 2020.

\bibitem[John et~al.(2018)John, Mou, Bahuleyan, and
  Vechtomova]{john-etal-2018-disentangled}
John, V., Mou, L., Bahuleyan, H., and Vechtomova, O.
\newblock Disentangled representation learning for non-parallel text style
  transfer.
\newblock \emph{arXiv preprint arXiv:1808.04339}, 2018.

\bibitem[Kingma \& Ba(2014)Kingma and Ba]{kingma-etal-2014-adam}
Kingma, D.~P. and Ba, J.
\newblock Adam: A method for stochastic optimization.
\newblock \emph{arXiv preprint arXiv:1412.6980}, 2014.

\bibitem[Kingma \& Welling(2013)Kingma and Welling]{kingma-etal-2013-auto}
Kingma, D.~P. and Welling, M.
\newblock Auto-encoding variational bayes.
\newblock \emph{arXiv preprint arXiv:1312.6114}, 2013.

\bibitem[Kingma et~al.(2014)Kingma, Mohamed, Rezende, and
  Welling]{kingma-etal-2014-semi}
Kingma, D.~P., Mohamed, S., Rezende, D.~J., and Welling, M.
\newblock Semi-supervised learning with deep generative models.
\newblock In \emph{Advances in neural information processing systems}, pp.\
  3581--3589, 2014.

\bibitem[Lample et~al.(2018)Lample, Subramanian, Smith, Denoyer, Ranzato, and
  Boureau]{lample-etal-2018-multiple}
Lample, G., Subramanian, S., Smith, E., Denoyer, L., Ranzato, M., and Boureau,
  Y.-L.
\newblock Multiple-attribute text rewriting.
\newblock In \emph{International Conference on Learning Representations}, 2018.

\bibitem[Lei et~al.(2016)Lei, Barzilay, and
  Jaakkola]{lei-etal-2016-rationalizing}
Lei, T., Barzilay, R., and Jaakkola, T.
\newblock Rationalizing neural predictions.
\newblock \emph{arXiv preprint arXiv:1606.04155}, 2016.

\bibitem[Lewis et~al.(2019)Lewis, Liu, Goyal, Ghazvininejad, Mohamed, Levy,
  Stoyanov, and Zettlemoyer]{lewis-etal-2019-bart}
Lewis, M., Liu, Y., Goyal, N., Ghazvininejad, M., Mohamed, A., Levy, O.,
  Stoyanov, V., and Zettlemoyer, L.
\newblock Bart: Denoising sequence-to-sequence pre-training for natural
  language generation, translation, and comprehension.
\newblock \emph{arXiv preprint arXiv:1910.13461}, 2019.

\bibitem[Li et~al.(2018)Li, Jia, He, and Liang]{li-etal-2018-delete}
Li, J., Jia, R., He, H., and Liang, P.
\newblock Delete, retrieve, generate: A simple approach to sentiment and style
  transfer.
\newblock \emph{arXiv preprint arXiv:1804.06437}, 2018.

\bibitem[Logeswaran et~al.(2018)Logeswaran, Lee, and
  Bengio]{logeswaran-etal-2018-content}
Logeswaran, L., Lee, H., and Bengio, S.
\newblock Content preserving text generation with attribute controls.
\newblock \emph{Advances in Neural Information Processing Systems}, 31, 2018.

\bibitem[Luo et~al.(2019)Luo, Li, Zhou, Yang, Chang, Sui, and
  Sun]{luo-etal-2019-dual}
Luo, F., Li, P., Zhou, J., Yang, P., Chang, B., Sui, Z., and Sun, X.
\newblock A dual reinforcement learning framework for unsupervised text style
  transfer.
\newblock \emph{arXiv preprint arXiv:1905.10060}, 2019.

\bibitem[Maal{\o}e et~al.(2016)Maal{\o}e, S{\o}nderby, S{\o}nderby, and
  Winther]{maaloe-etal-2016-auxiliary}
Maal{\o}e, L., S{\o}nderby, C.~K., S{\o}nderby, S.~K., and Winther, O.
\newblock Auxiliary deep generative models.
\newblock In \emph{International conference on machine learning}, pp.\
  1445--1453. PMLR, 2016.

\bibitem[Madaan et~al.(2020)Madaan, Setlur, Parekh, Poczos, Neubig, Yang,
  Salakhutdinov, Black, and Prabhumoye]{madaan-etal-2020-politeness}
Madaan, A., Setlur, A., Parekh, T., Poczos, B., Neubig, G., Yang, Y.,
  Salakhutdinov, R., Black, A.~W., and Prabhumoye, S.
\newblock Politeness transfer: A tag and generate approach.
\newblock \emph{arXiv preprint arXiv:2004.14257}, 2020.

\bibitem[Makhzani et~al.(2015)Makhzani, Shlens, Jaitly, Goodfellow, and
  Frey]{makhzani-etal-2015-adversarial}
Makhzani, A., Shlens, J., Jaitly, N., Goodfellow, I., and Frey, B.
\newblock Adversarial autoencoders.
\newblock \emph{arXiv preprint arXiv:1511.05644}, 2015.

\bibitem[Mescheder et~al.(2017)Mescheder, Nowozin, and
  Geiger]{mescheder-etal-2017-adversarial}
Mescheder, L., Nowozin, S., and Geiger, A.
\newblock Adversarial variational bayes: Unifying variational autoencoders and
  generative adversarial networks.
\newblock In \emph{International Conference on Machine Learning}, pp.\
  2391--2400. PMLR, 2017.

\bibitem[Mohamed \& Lakshminarayanan(2016)Mohamed and
  Lakshminarayanan]{mohamed-etal-2016-learning}
Mohamed, S. and Lakshminarayanan, B.
\newblock Learning in implicit generative models.
\newblock \emph{arXiv preprint arXiv:1610.03483}, 2016.

\bibitem[Molnar(2020)]{molnar-etal-2020-interpretable}
Molnar, C.
\newblock \emph{Interpretable machine learning}.
\newblock Lulu. com, 2020.

\bibitem[Nangi et~al.(2021)Nangi, Chhaya, Khosla, Kaushik, and
  Nyati]{nangi-etal-2021-counterfactuals}
Nangi, S.~R., Chhaya, N., Khosla, S., Kaushik, N., and Nyati, H.
\newblock Counterfactuals to control latent disentangled text representations
  for style transfer.
\newblock In \emph{Proceedings of the 59th Annual Meeting of the Association
  for Computational Linguistics and the 11th International Joint Conference on
  Natural Language Processing (Volume 2: Short Papers)}, pp.\  40--48, 2021.

\bibitem[Nowozin et~al.(2016)Nowozin, Cseke, and Tomioka]{nowozin-etal-2016-f}
Nowozin, S., Cseke, B., and Tomioka, R.
\newblock f-gan: Training generative neural samplers using variational
  divergence minimization.
\newblock In \emph{Proceedings of the 30th International Conference on Neural
  Information Processing Systems}, pp.\  271--279, 2016.

\bibitem[Papineni et~al.(2002)Papineni, Roukos, Ward, and
  Zhu]{papineni-etal-2002-bleu}
Papineni, K., Roukos, S., Ward, T., and Zhu, W.-J.
\newblock Bleu: a method for automatic evaluation of machine translation.
\newblock In \emph{Proceedings of the 40th annual meeting of the Association
  for Computational Linguistics}, pp.\  311--318, 2002.

\bibitem[Pennington et~al.(2014)Pennington, Socher, and
  Manning]{pennington-etal-2014-glove}
Pennington, J., Socher, R., and Manning, C.~D.
\newblock Glove: Global vectors for word representation.
\newblock In \emph{Proceedings of the 2014 conference on empirical methods in
  natural language processing (EMNLP)}, pp.\  1532--1543, 2014.

\bibitem[Prabhumoye et~al.(2018)Prabhumoye, Tsvetkov, Salakhutdinov, and
  Black]{prabhumoye-etal-2018-style}
Prabhumoye, S., Tsvetkov, Y., Salakhutdinov, R., and Black, A.~W.
\newblock Style transfer through back-translation.
\newblock \emph{arXiv preprint arXiv:1804.09000}, 2018.

\bibitem[Radford et~al.(2019)Radford, Wu, Child, Luan, Amodei, Sutskever,
  et~al.]{radford-etal-2019-language}
Radford, A., Wu, J., Child, R., Luan, D., Amodei, D., Sutskever, I., et~al.
\newblock Language models are unsupervised multitask learners.
\newblock \emph{OpenAI blog}, 1\penalty0 (8):\penalty0 9, 2019.

\bibitem[Rezende et~al.(2014)Rezende, Mohamed, and
  Wierstra]{rezende-etal-2014-stochastic}
Rezende, D.~J., Mohamed, S., and Wierstra, D.
\newblock Stochastic backpropagation and variational inference in deep latent
  gaussian models.
\newblock In \emph{International Conference on Machine Learning}, volume~2,
  pp.\ ~2. Citeseer, 2014.

\bibitem[Ribeiro et~al.(2016)Ribeiro, Singh, and
  Guestrin]{ribeiro-etal-2016-should}
Ribeiro, M.~T., Singh, S., and Guestrin, C.
\newblock " why should i trust you?" explaining the predictions of any
  classifier.
\newblock In \emph{Proceedings of the 22nd ACM SIGKDD international conference
  on knowledge discovery and data mining}, pp.\  1135--1144, 2016.

\bibitem[Romanov et~al.(2018)Romanov, Rumshisky, Rogers, and
  Donahue]{romanov-etal-2018-adversarial}
Romanov, A., Rumshisky, A., Rogers, A., and Donahue, D.
\newblock Adversarial decomposition of text representation.
\newblock \emph{arXiv preprint arXiv:1808.09042}, 2018.

\bibitem[Serban et~al.(2016)Serban, Sordoni, Bengio, Courville, and
  Pineau]{serban-etal-2016-building}
Serban, I., Sordoni, A., Bengio, Y., Courville, A., and Pineau, J.
\newblock Building end-to-end dialogue systems using generative hierarchical
  neural network models.
\newblock In \emph{Proceedings of the AAAI Conference on Artificial
  Intelligence}, volume~30, 2016.

\bibitem[Serban et~al.(2017)Serban, Sordoni, Lowe, Charlin, Pineau, Courville,
  and Bengio]{serban-etal-2017-hierarchical}
Serban, I., Sordoni, A., Lowe, R., Charlin, L., Pineau, J., Courville, A., and
  Bengio, Y.
\newblock A hierarchical latent variable encoder-decoder model for generating
  dialogues.
\newblock In \emph{Proceedings of the AAAI Conference on Artificial
  Intelligence}, volume~31, 2017.

\bibitem[Serrano \& Smith(2019)Serrano and Smith]{serrano-etal-2019-attention}
Serrano, S. and Smith, N.~A.
\newblock Is attention interpretable?
\newblock \emph{arXiv preprint arXiv:1906.03731}, 2019.

\bibitem[Shang et~al.(2019)Shang, Li, Fu, Bing, Zhao, Shi, and
  Yan]{shang-etal-2019-semi}
Shang, M., Li, P., Fu, Z., Bing, L., Zhao, D., Shi, S., and Yan, R.
\newblock Semi-supervised text style transfer: Cross projection in latent
  space.
\newblock \emph{arXiv preprint arXiv:1909.11493}, 2019.

\bibitem[Shen et~al.(2017)Shen, Lei, Barzilay, and
  Jaakkola]{shen-etal-2017-style}
Shen, T., Lei, T., Barzilay, R., and Jaakkola, T.
\newblock Style transfer from non-parallel text by cross-alignment.
\newblock \emph{arXiv preprint arXiv:1705.09655}, 2017.

\bibitem[Shen et~al.(2020)Shen, Quach, Barzilay, and
  Jaakkola]{shen-etal-2020-blank}
Shen, T., Quach, V., Barzilay, R., and Jaakkola, T.
\newblock Blank language models.
\newblock \emph{arXiv preprint arXiv:2002.03079}, 2020.

\bibitem[Simonyan et~al.(2013)Simonyan, Vedaldi, and
  Zisserman]{simonyan-etal-2013-deep}
Simonyan, K., Vedaldi, A., and Zisserman, A.
\newblock Deep inside convolutional networks: Visualising image classification
  models and saliency maps.
\newblock \emph{arXiv preprint arXiv:1312.6034}, 2013.

\bibitem[Smith(2017)]{smith-etal-2017-cyclical}
Smith, L.~N.
\newblock Cyclical learning rates for training neural networks.
\newblock In \emph{2017 IEEE winter conference on applications of computer
  vision (WACV)}, pp.\  464--472. IEEE, 2017.

\bibitem[Stolcke(2002)]{stolcke-etal-2002-srilm}
Stolcke, A.
\newblock Srilm-an extensible language modeling toolkit.
\newblock In \emph{Seventh international conference on spoken language
  processing}, 2002.

\bibitem[Sudhakar et~al.(2019)Sudhakar, Upadhyay, and
  Maheswaran]{sudhakar-etal-2019-transforming}
Sudhakar, A., Upadhyay, B., and Maheswaran, A.
\newblock Transforming delete, retrieve, generate approach for controlled text
  style transfer.
\newblock \emph{arXiv preprint arXiv:1908.09368}, 2019.

\bibitem[Sundararajan et~al.(2017)Sundararajan, Taly, and
  Yan]{sundararajan-etal-2017-axiomatic}
Sundararajan, M., Taly, A., and Yan, Q.
\newblock Axiomatic attribution for deep networks.
\newblock In \emph{International Conference on Machine Learning}, pp.\
  3319--3328. PMLR, 2017.

\bibitem[Tomczak \& Welling(2018)Tomczak and Welling]{tomczak-etal-2018-vae}
Tomczak, J. and Welling, M.
\newblock Vae with a vampprior.
\newblock In \emph{International Conference on Artificial Intelligence and
  Statistics}, pp.\  1214--1223. PMLR, 2018.

\bibitem[Van~der Maaten \& Hinton(2008)Van~der Maaten and
  Hinton]{van-etal-200-8visualizing}
Van~der Maaten, L. and Hinton, G.
\newblock Visualizing data using t-sne.
\newblock \emph{Journal of machine learning research}, 9\penalty0 (11), 2008.

\bibitem[Vaswani et~al.(2017)Vaswani, Shazeer, Parmar, Uszkoreit, Jones, Gomez,
  Kaiser, and Polosukhin]{vaswani-etal-2017-attention}
Vaswani, A., Shazeer, N., Parmar, N., Uszkoreit, J., Jones, L., Gomez, A.~N.,
  Kaiser, {\L}., and Polosukhin, I.
\newblock Attention is all you need.
\newblock In \emph{Advances in neural information processing systems}, pp.\
  5998--6008, 2017.

\bibitem[Wang et~al.(2019)Wang, Hua, and Wan]{wang-etal-2019-controllable}
Wang, K., Hua, H., and Wan, X.
\newblock Controllable unsupervised text attribute transfer via editing
  entangled latent representation.
\newblock \emph{Advances in Neural Information Processing Systems},
  32:\penalty0 11036--11046, 2019.

\bibitem[Xu et~al.(2018)Xu, Sun, Zeng, Ren, Zhang, Wang, and
  Li]{xu-etal-2018-unpaired}
Xu, J., Sun, X., Zeng, Q., Ren, X., Zhang, X., Wang, H., and Li, W.
\newblock Unpaired sentiment-to-sentiment translation: A cycled reinforcement
  learning approach.
\newblock \emph{arXiv preprint arXiv:1805.05181}, 2018.

\bibitem[Yang et~al.(2018)Yang, Hu, Dyer, Xing, and
  Berg-Kirkpatrick]{yang-etal-2018-unsupervised}
Yang, Z., Hu, Z., Dyer, C., Xing, E.~P., and Berg-Kirkpatrick, T.
\newblock Unsupervised text style transfer using language models as
  discriminators.
\newblock In \emph{Proceedings of the 32nd International Conference on Neural
  Information Processing Systems}, pp.\  7298--7309, 2018.

\bibitem[Zhang et~al.(2018{\natexlab{a}})Zhang, Xu, Yang, and
  Sun]{zhang-etal-2018-learning}
Zhang, Y., Xu, J., Yang, P., and Sun, X.
\newblock Learning sentiment memories for sentiment modification without
  parallel data.
\newblock \emph{arXiv preprint arXiv:1808.07311}, 2018{\natexlab{a}}.

\bibitem[Zhang et~al.(2018{\natexlab{b}})Zhang, Ren, Liu, Wang, Chen, Li, Zhou,
  and Chen]{zhang-etal-2018-style}
Zhang, Z., Ren, S., Liu, S., Wang, J., Chen, P., Li, M., Zhou, M., and Chen, E.
\newblock Style transfer as unsupervised machine translation.
\newblock \emph{arXiv preprint arXiv:1808.07894}, 2018{\natexlab{b}}.

\bibitem[Zhao et~al.(2017)Zhao, Song, and Ermon]{zhao-etal-2017-infovae}
Zhao, S., Song, J., and Ermon, S.
\newblock Infovae: Information maximizing variational autoencoders.
\newblock \emph{arXiv preprint arXiv:1706.02262}, 2017.

\end{thebibliography}

\appendix

\section{Derivation Details}
\subsection{ELBO Decomposition (Equation (10)) in Section 4}
\label{app:elbo_decompose}
In this formulation, only the KL-regularization term is decomposed. According to its definition, we have
\begin{equation}
{\Bbb E}_{p_{\cal D}(x,y)}{\rm KL}(q_\phi(c|x)\Vert p(c))={\Bbb E}_{p_{\cal D}(y)}\left[\iint p_{\cal D}(x|y)q_\phi(c|x)\log\frac{q_\phi(c|x)}{p(c)}dcdx\right]\\
\end{equation}
Since we assume $q_\phi(c|x)=q_\phi(c|x,y)$, we let $\tilde{q}_\phi(c,x|y)=p_{\cal D}(x|y)q_\phi(c|x)$, and $\tilde{q}_\phi(c|y)=\int\tilde{q}_\phi(c,x|y)dx$ as the aggregated posterior of content code $c$ with style $y$. Then we have
\begin{equation}
\begin{aligned}
\iint p_{\cal D}(x|y)q_\phi(c|x)&\log\frac{q_\phi(c|x)}{p(c)}dcdx=\iint\tilde{q}_\phi(c,x|y)\log\frac{q_\phi(c|x)}{p(c)}dcdx\\
&=\iint\tilde{q}_\phi(c,x|y)\log\frac{q_\phi(c|x)\tilde{q}_\phi(c|y)}{\tilde{q}_\phi(c|y)p(c)}dcdx\\
&=\int\tilde{q}_\phi(c|y)\log\frac{\tilde{q}_\phi(c|y)}{p(c)}dc+\iint\tilde{q}_\phi(c,x|y)\log\frac{\tilde{q}_\phi(c,x|y)}{\tilde{q}_\phi(c|y)p_D(x|y)}dcdx\\
&={\rm KL}(\tilde{q}_\phi(c|y)\Vert p(c))+I_{\tilde{q}(c,x|y)}(x;c)
\end{aligned}
\end{equation}
Thus,
\begin{equation}
{\Bbb E}_{p_{\cal D}(x,y)}{\rm KL}(q_\phi(c|x)\Vert p(c))={\Bbb E}_{p_{\cal D}(y)}\left[{\rm KL}(\tilde{q}_\phi(c|y)\Vert p(c))+I_{\tilde{q}(c,x|y)}(x;c)\right]
\end{equation}
Further, ${\Bbb E}_{p_{\cal D}(y)}\left[{\rm KL}(\tilde{q}_\phi(c|y)\Vert p(c))\right]$ is decomposed as:
\begin{equation}
\begin{aligned}
{\Bbb E}_{p_{\cal D}(y)}&\left[\text{KL}(\tilde{q}_\phi(c|y)\Vert p(c))\right]=\iint p_{\cal D}(y)\tilde{q}_\phi(c|y)\log\frac{\tilde{q}_\phi(c|y)}{p(c)}\\
&=\iint p_{\cal D}(y)\tilde{q}_\phi(c|y)\log\tilde{q}_\phi(c|y)dcdy-\int\tilde{q}_\phi(c)\log p(c)dc\\
&=\iint p_{\cal D}(y)\tilde{q}_\phi(c|y)\log\tilde{q}_\phi(c|y)dcdy-\iint p_{\cal D}(y)\tilde{q}_\phi(c|y)\log \tilde{q}_\phi(c)dcdy\\
&\quad + \int\tilde{q}_\phi(c)\log \tilde{q}_\phi(c)dc - \int\tilde{q}_\phi(c)\log p(c)dc\\
&=\iint p_{\cal D}(y)\tilde{q}_\phi(c|y)\log\frac{\tilde{q}_\phi(y|c)}{\int p_{\cal D}(y')\tilde{q}_\phi(c|y')dy'}dcdy+\int\tilde{q}_\phi(c)\log\frac{\tilde{q}_\phi(c)}{p(c)}dc\\
&=\text{JS}(\Vert_y\tilde{q}_\phi(c|y))+\text{KL}(\tilde{q}_\phi(c)\Vert p(c))
\end{aligned}
\end{equation}
Then combining with the reconstruction loss would lead to Equation (10).

\subsection{Objective of Prototype Form (Equation (17)) in Section 5.2}
\label{app:mask_total_loss}
We consider ELBO loss in Equation (5) and classification loss in Equation (8) individually. ELBO loss contains reconstruction loss and KL-regularization loss. 
According to the property of delta function, the former becomes:
\begin{equation}
\begin{aligned}
-{\Bbb E}_{q_\phi(c|x)}[\log p_\theta(x|c,y)]&=-\log p_\theta(x|c_\phi(x),y)
\end{aligned}    
\end{equation}
For the latter we have
\begin{equation}
\begin{aligned}
\text{KL}(q_\phi(c|x)\Vert p(c))&=\text{KL}(q_\phi(\bar{c}|x)\Vert p(\bar{c}))=-\log p(\bar{c}_\phi(x))\\
&=-\sum_{t=1}^{|x|}\log p(\bar{c}_\phi^{(t)}(x))-\log p(|x|)\\
&=-\sum_{t=1}^{|x|}\left[\log r_{|x|}\cdot \bar{c}_\phi^{(t)}(x)+\log (1-r_{|x|})\cdot(1-\bar{c}_\phi^{(t)}(x))\right]-\log p(|x|)\\
&=\log\left(\frac{1-r_{|x|}}{r_{|x|}}\right)\sum_{t=1}^{|x|}\bar{c}_\phi^{(t)}(x)-\log p(|x|)-|x|\log (1-r_{|x|})
\end{aligned}
\end{equation}
By substituting $\bar{c}_\phi^{(t)}(x)$ with $1-\bar{s}_\phi^{(t)}(x)$, it gives:
\begin{equation}
\text{KL}(q_\phi(c|x)\Vert p(c))=\log\left(\frac{r_{|x|}}{1-r_{|x|}}\right)\sum_{t=1}^{|x|}\bar{s}_\phi^{(t)}(x)+\text{const.}
\end{equation}

For classification loss, analogous to reconstruction loss, we have
\begin{equation}
    -{\Bbb E}_{q_\phi(c|x)q_\phi(s|x,c)}[\log q_\phi(y|s)]=-\log q_\phi(y|s_\phi(x))
\end{equation}

Then simply combining them together and omitting constant terms would lead to Equation (17).

\begin{figure}
    \centering
    \begin{subfigure}[b]{0.22\linewidth}
        \centering
        \includegraphics[width=\linewidth]{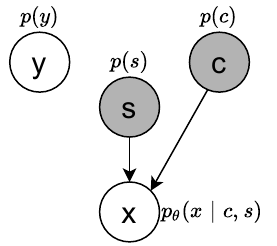}
        \caption{Generative Model ${\Bbb P}$.}
        \label{fig:generative_pgm}
    \end{subfigure}
    \begin{subfigure}[b]{0.22\linewidth}
        \centering
        \includegraphics[width=\linewidth]{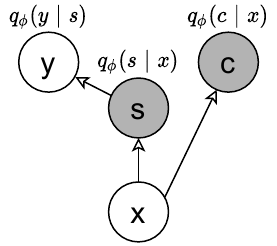}
        \caption{Inference Model ${\Bbb Q}$.}
        \label{fig:inference_pgm}
    \end{subfigure}
    \caption{Probabilistic graph of RS framework, where shadowed nodes represent latent variables.}
    \label{fig:rs_pgm}
\end{figure}

\section{Learning of Embedding Form Method}
We drop the MI term in Equation (10) to avoid posterior collapse and the resulting objective is (Equation (12)):
\begin{equation}
{\cal L}_{\theta,\phi}^{\rm ELBO'}={\Bbb E}_{p_{\cal D}(y)}\left[-{\Bbb E}_{p_{\cal D}(x|y)}[\log p_\theta(x|c_\phi(x),y)]+\text{KL}(\tilde{q}_\phi(c|y)\Vert p(c))\right]
\end{equation}
The second KL term is unable to be computed analytically. Note that this term is the expectation of a density ratio:
\begin{equation}
{\Bbb E}_{p_{\cal D}(y)}\text{KL}(\tilde{q}(c|y)\Vert p(c))={\Bbb E}_{p_{\cal D}(x,y)q_\phi(c|x)}[\log\tilde{q}_\phi(c|y)-\log p(c)]
\end{equation}
The density ratio could be estimated with a discriminator $f_{\omega_y}^\star(c)$ \citep{mescheder-etal-2017-adversarial}, which comes from the following optimization problem:
\begin{equation}
\omega_y^\star=\mathop{\arg\max}_{\omega_y}{\Bbb E}_{p_{\cal D}(x,y)q_\phi(c|x)}\big[\log\sigma(f_{\omega_y}(c))\big]+{\Bbb E}_{p_{\cal D}(y)p(c)}\big[\log(1-\sigma(f_{\omega_y}(c)))\big]
\end{equation}
It is proved by \citet{mescheder-etal-2017-adversarial} that the optimal discriminator satisfies $f_{\omega_y}^\star(c)=\log\tilde{q}_\phi(c|y)-\log p(c)$. Therefore the KL divergence becomes:
\begin{equation}
{\Bbb E}_{p_{\cal D}(y)}[\text{KL}(\tilde{q}_\phi(c|y)\Vert p(c))]={\Bbb E}_{p_{\cal D}(x,y)q_\phi(c|x)}[f_{w_y}^\star(c)]
={\Bbb E}_{p_{\cal D}(x,y)}[f_{w_y}^\star(c_\phi(x))]
\end{equation}
The training procedure is alternated analogous to generative adversarial networks (GANs) \cite{goodfellow-etal-2014-generative}. We parameterize the discriminator using multilayer perceptrons.

\section{Representation Splitting Framework}
\label{app:rs}
The generative model and inference model of this framework are shown in Figure \ref{fig:rs_pgm}, which we have:
\begin{equation}
\begin{aligned}
p_\theta(x,y,s,c)&=p(y)p(s)p(c)p_\theta(x|s,c)\\
q_\phi(y,s,c|x)&=q_\phi(c|x)q_\phi(s|x)q_\phi(y|s)
\end{aligned}
\end{equation}
As seen, There are some major difference between our framework and RS framework:
\begin{description}[style=unboxed,leftmargin=0.2cm]

\item[-] For generative model, firstly, the RS framework doesn't consider how the style label $y$ affects the generation of sentence $x$. Therefore its first goal $p_\theta(x|y)$ degenerates to only the marginal distribution $p_\theta(x)$ such that the dependency of $x$ on $y$ is not exploited. Since we've established in Section 4 that much power of a disentanglement-based system comes from aligning contents across styles, this architecture is clearly not desired. Secondly, the RS framework treats $s$ and $c$ as two independent generative factors of $x$, which is irrational because stylish expressions (represented by $s$) usually have a strong relatedness or dependence with the content pieces (represented by $c$), such as ``delicious" for ``food" and ``patient" for ``service". This shortcoming is further exacerbated such that many works arbitrarily assume the prior $p(s)$ as a uni-modal Gaussian \citep{john-etal-2018-disentangled, bao-etal-2019-generating, nangi-etal-2021-counterfactuals}, where it's factually more appropriate to be multi-modal. In contrast, our generative model is plausible and avoids setting this complicated prior. Specifically, in our approach $p_\theta(x,y,c)$ is invariant to marginalization over $s$: $p_\theta(x,y,c,s)=p_\theta(x,y,c)p(s|x)$, which significantly simplifies the problem such that $s$ is out of consider in modeling $p(x|y)$ as we've seen.

\item[-] For inference model, the RS framework applies mean-field assumptions of the encoder $q_\phi(s,c|x)=q_\phi(c|x)q_\phi(s|x)$, but our framework expand this completely $q_\phi(s,c|x)=q_\phi(c|x)q_\phi(s|x,c)$. While this form is more general, the term $q_\phi(s|x,c)$ entails an intuition that the style code $s$ could be inferred by ``subtracting'' content code $c$ from the sentence $x$. In this way $c$ and $s$ are connected and can be enhanced by exploiting each other. Furthermore, we show in Section 5.2 that this assumption is the key to unify prototype form methods.

\end{description}

Analogy to our framework, the first goal of this framework is to maximize the logarithm likelihood $p(x)$:
\begin{equation}
\log p_\theta(x)\geq{\Bbb E}_{q_\phi(c|x)q_\phi(s|x)}[\log p_\theta(x|c,s)] - \text{KL}(q_\phi(c|x)\Vert p(c)) - \text{KL}(q_\phi(s|x)\Vert p(s))\equiv-{\cal L}_{\theta,\phi}^\text{RS\_ELBO}
\end{equation}
The second goal is to learn the variational classifier $q_\phi(y|x)$ by maximizing:
\begin{equation}
\log q_\phi(y|x)=\log {\Bbb E}_{q_\phi(s|x)}[q_\phi(y|s)]\geq{\Bbb E}_{q_\phi(s|x)}[q_\phi(y|s)]\equiv-{\cal L}_{\phi}^\text{RS\_CLS}
\end{equation}
The total loss of this method is therefore a linear combination of these two losses:
\begin{equation}
{\cal L}_{\theta,\phi}^\text{RS}={\Bbb E}_{p_{\cal D}(x,y)}[{\cal L}_{\theta,\phi}^\text{RS\_ELBO}+\gamma{\cal L}_{\phi}^\text{RS\_CLS}]
\end{equation}
Note that this model is unable to generalize to the prototype form since $c$ and $s$ are conditional independent when $x$ is known. So we only experiment with embedding form, which assumes $p(c)$ and $p(s)$ as unit Gaussian distributions following previous works. The ELBO term in it is also decomposed as:
\begin{equation}
\begin{aligned}
{\cal L}_{\theta,\phi}^\text{RS\_ELBO}=\underbrace{-{\Bbb E}_{p_{\cal D}(x)q_\phi(c|x)q_\phi(s|x)}[\log p_\theta(x|s,c)]+\text{KL}(\tilde{q}_\phi(c)\Vert p(c))+\text{KL}(\tilde{q}_\phi(s)\Vert p(s))}_{{\cal L}_{\theta,\phi}^\text{RS\_ELBO'}}\\
+I_{\tilde{q}_\phi(x,s)}(x;s)+I_{\tilde{q}_\phi(x,c)}(x;c)
\end{aligned}
\end{equation}
where
\begin{equation}
\begin{aligned}
\tilde{q}_\phi(x,c)&=q_\phi(c|x)p_{\cal D}(x),\quad\tilde{q}(c)={\Bbb E}_{p_{\cal D}(x)}[q_\phi(c|x)]\\
\tilde{q}_\phi(x,s)&=q_\phi(s|x)p_{\cal D}(x),\quad\tilde{q}(s)={\Bbb E}_{p_{\cal D}(x)}[q_\phi(s|x)]
\end{aligned}
\end{equation}
We also drop MI terms and only optimize ${\cal L}_{\theta,\phi}^\text{RS\_ELBO'}$ for fair comparison. The method also follows Equation (2) in Section 2.2.1, which further defines the decoder as
\begin{equation}
p(x'|c,y')=\int p_\theta(x'|c,s')\tilde{q}_\phi(s'|y')ds'=\iint p_\theta(x'|c,s')q_\phi(s'|x'')p_{\cal D}(x''|y')dx''ds'
\end{equation}
When the training process is over, the transferred sentence is obtained by:
\begin{equation}
x'^\star=\mathop{\text{BeamSearch}}_{x'} p_\theta(x'|{\Bbb E}_{q_\phi(c|x)}[c],{\Bbb E}_{p_{\cal D}(x''|y')q_\phi(s'|x'')}[s'])
\end{equation}

\section{Model Architecture \& Hyperparameters}
\label{app:hyperparmeters}
For embedding form methods including baselines, we use same structures of both encoder, decoder and discriminators on each dataset, as shown in Table \ref{tab:emb_hyperparameters}. GloVe embeddings \citep{pennington-etal-2014-glove} can be found in \url{https://nlp.stanford.edu/projects/glove/}. All the models are optimized using Adam \citep{kingma-etal-2014-adam}. We set the learning rate to be $1\times 10^{-3}$ for the encoder and decoder, $5\times 10^{-2}$ for the discriminator. We update the encoder and decoder once after five updates of the discriminator.

For prototype form methods including baselines, we share the same decoder structure. Hyperparameters are shown in Table \ref{tab:prototype_hyperparameters}. We also use Adam algorithm for optimization and set the learning to be $5\times 10^{-4}$, $1\times 10^{-4}$, $1\times 10^{-3}$ for \textsc{Yelp}, \textsc{Amazon} and \textsc{Captions} respectively when minimizing ${\cal L}_\phi^\text{RAT}$. When we optimize ${\cal L}_{\theta,\phi^\star}^\text{INFILL}$, we utilize cyclic learning rate technique \citep{smith-etal-2017-cyclical}. Base learning rate and maximize learning rate are set to $1\times 10^{-4}$ and $1\times 10^{-3}$, respectively for all datasets. and the cyclic step interval is set to 6000 for \textsc{Yelp} and \textsc{Amazon}, and 200 for \textsc{Captions}.

The parameter size of our method is not large. Each run is conduced on a GeForce RTX 2080Ti chip.

\begin{table*}[t]
\centering
\caption{Model hyerparmeters of embedding form methods, where $d_h$, $d_c$ and $d_s$ stand for the dimension of hidden states in LSTM, content code and style code respectively.}
\label{tab:emb_hyperparameters}
\begin{tabular}{lcccccccc}
\toprule
\multirow{2}{*}{Dataset} & \multirow{2}{*}{$\gamma$} & \multirow{2}{*}{Embedding} & \multicolumn{3}{c}{\makecell[c]{Encoder\\BiLSTM}}  & \makecell[c]{Decoder\\LSTM} & \multicolumn{2}{c}{\makecell[c]{Discriminator\\MLP}} \\ \cmidrule(r){4-6}\cmidrule(r){7-7}\cmidrule(r){8-9} 
                         &                        &                            & $d_h$        & $d_c$ & $d_s$ & $d_h$   & Dims of layers    & Activation    \\ \midrule
\textsc{Yelp}                     & 1                      & GloVe 300d                 & $256\times2$ & 26    & 26    & 256     & 26, 52, 1         & ReLU          \\
\textsc{Amazon}                   & 1                      & GloVe 300d                 & $256\times2$ & 20    & 20    & 512     & 20, 40, 1         & ReLU          \\
\textsc{Captions}                 & 1                      & GloVe 300d                 & $256\times2$ & 32    & 32    & 256     & 32, 64, 1         & ReLU          \\ \bottomrule
\end{tabular}
\end{table*}

\begin{table*}[t]
\centering
\caption{Model hyerparmeters of prototype form methods, where $d_h$ stands for the dimension of hidden states in LSTM.}
\label{tab:prototype_hyperparameters}
\begin{tabular}{lcccccc}
\toprule
\multirow{2}{*}{Dataset} & \multirow{2}{*}{$\gamma$} & \multirow{2}{*}{$\alpha$} & \multirow{2}{*}{Embedding} & \makecell[c]{Encoder\\BiLSTM}     & \makecell[c]{Classifier\\BiLSTM}  & \makecell[c]{Decoder\\seq2seq LSTM + attn} \\ \cmidrule(r){5-5}\cmidrule(r){6-6}\cmidrule(r){7-7}
                         &                        &                        &                            & $d_h$        & $d_h$        & $d_h$   \\ \hline
\textsc{Yelp}                     & 1                      & 0.150                  & GloVe 300d                 & $256\times2$ & $256\times2$ & 256     \\
\textsc{Amazon}                   & 1                      & 0.190                  & GloVe 300d                 & $384\times2$ & $384\times2$ & 768     \\
\textsc{Captions}                 & 1                      & 0.225                  & GloVe 300d                 & $256\times2$ & $256\times2$ & 256     \\ \bottomrule
\end{tabular}
\end{table*}

\section{Disentanglement of Latent Vectors in Embedding Form}

\begin{figure}
    \centering
    \begin{subfigure}[b]{0.25\linewidth}
        \centering
        \includegraphics[width=\linewidth]{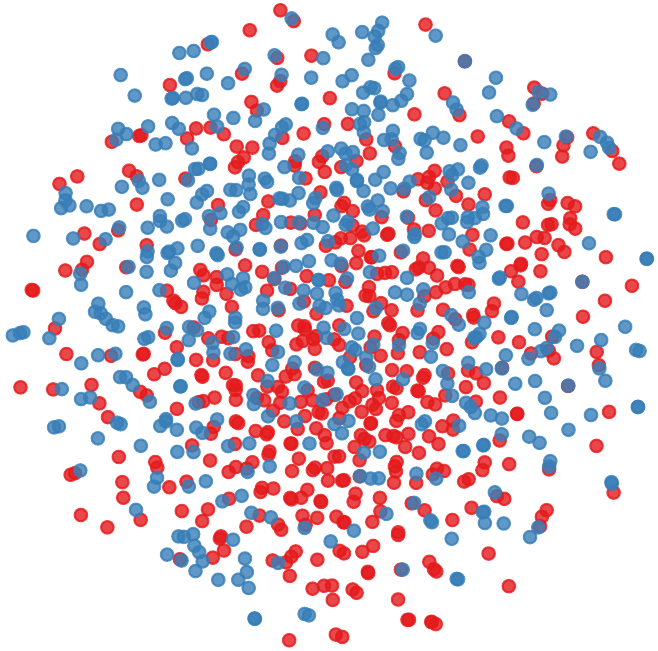}
        \caption{Content space.}
    \end{subfigure}
    \begin{subfigure}[b]{0.25\linewidth}
        \centering
        \includegraphics[width=\linewidth]{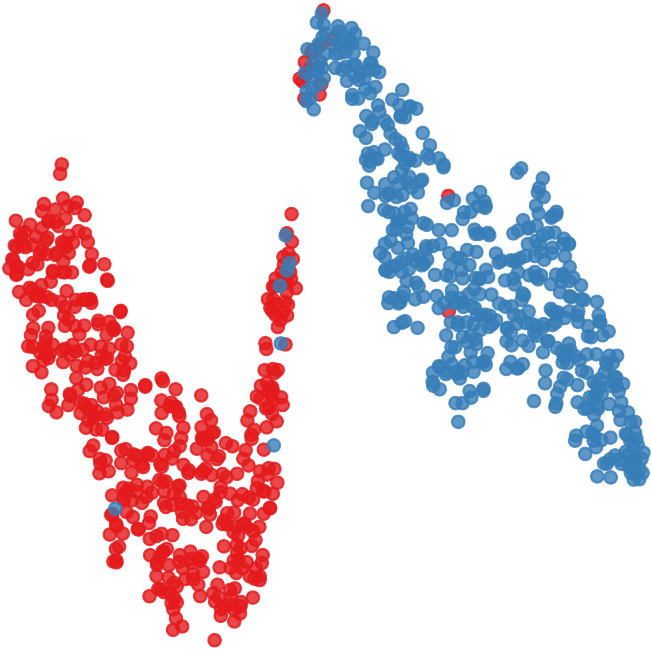}
        \caption{Style space.}
    \end{subfigure}
    \caption{Visualizations of latent codes on \textsc{Yelp}. Blue/Red represents positive/negative sentences.}
    \label{fig:visual}
\end{figure}

To illustrate that latent vectors $s$ and $c$ in our embedding form method are truly disentangled, we train a linear logistic classifier on testing set of each dataset and report the classification accuracy in Table \ref{tab:disentanglement_accuracy}. As seen, the content vector $c$ is not discriminative to the styles, while its accuracy is slightly higher than majority guess in both three datasets. At the same time, the style vector $s$ is able to predict with high accuracy. These results verify that our method disentangles content and style well. We also visualize these vectors of sentences in \textsc{Yelp} using t-SNE \citep{van-etal-200-8visualizing} as shown in Figure \ref{fig:visual}.

\begin{table}[]
\centering
\caption{Classification accuracy on latent vectors of our embedding form method.}
\label{tab:disentanglement_accuracy}
\begin{tabular}{lcccccc}
\toprule
\multirow{2}{*}{Latent Space} & \multicolumn{2}{c}{\textsc{Yelp}} & \multicolumn{2}{c}{\textsc{Amazon}} & \multicolumn{2}{c}{\textsc{Captions}} \\ \cmidrule(r){2-3} \cmidrule(r){4-5} \cmidrule(r){6-7}
                              & Train       & Test       & Train        & Test        & Train         & Test         \\ \midrule
Majority Guess                & 61.19       & 50.00      & 49.95        & 50.00       & 50.00         & /            \\
Content Space ($c$)           & 66.08       & 63.20      & 60.11        & 56.80       & 57.33         & /            \\
Style Space ($s$)             & 98.68       & 97.90      & 86.55        & 82.50       & 99.90         & /            \\ \bottomrule
\end{tabular}
\end{table}

\section{Human Evaluation}
We recruit to raters for human evaluation. Both raters are graduate students with NLP background. They are familiar with the goal of research. They do human evaluation in voluntary since our human evaluation samples are not very huge. Here, we present text of instructions given to participants for human evaluation:
\paragraph{Instruction of \textsc{Yelp} and \textsc{Amazon}} You are required to evaluate the quality of sentiment transferred sentences in terms of transfer strength, content preservation and language quality.
\begin{description}[style=unboxed,leftmargin=0.2cm]
\item[- \textnormal{Transfer strength:}] You need to judge if the sentiment polarity of generated sentence is flipped and score the extent from 1 to 5 point: If the sentiment is totally flipped, score 5 point; If the sentiment is remained as the same of the original sentence, score 1 point; If the sentiment is neutral or unable to judge, score 3 point.
\item[- \textnormal{Content preservation:}] You need to judge if the content of generated sentence is maintained and score the extent from 1 to 5 point. Higher score suggests better content preservation.
\item[- \textnormal{Language quality:}] You need to judge the quality of generated sentence in terms of grammar, fluency and so on. Higher score suggests better language quality.
\end{description}
Examples in Table \ref{tab:human_eval_example_sentiment} should help you to understand the standard roughly.

\begin{table}[]
\centering
\scriptsize
\caption{Human evaluation examples on sentiment transfer (\textsc{Yelp} and \textsc{Amazon}).}
\label{tab:human_eval_example_sentiment}
\begin{tabular}{llccc}
\toprule
Transfer Direction & Source (top) and transferred (bottom) sentences & \makecell[c]{Transfer\\Strength}  & \makecell[c]{Content\\Preservation} & \makecell[c]{Language\\Quality}   \\ \midrule
\multirow{4}{*}{\makecell[l]{From positive\\to negative}} & the biscuits and gravy were good .                      & \multirow{2}{*}{5} & \multirow{2}{*}{5}   & \multirow{2}{*}{5} \\
                                           & the biscuits and gravy were bland .                     &                    &                      &                    \\ \cmidrule{2-5} 
                                           & everyone that i spoke with was very helpful and kind .  & \multirow{2}{*}{3} & \multirow{2}{*}{5}   & \multirow{2}{*}{3} \\
                                           & everyone that i spoke with was very helpful and rude .  &                    &                      &                    \\ \midrule
\multirow{4}{*}{\makecell[l]{From negative\\to positive}} & the charge did include miso soup and a small salad .    & \multirow{2}{*}{5} & \multirow{2}{*}{2}   & \multirow{2}{*}{4} \\
                                           & the salads are paired with a slaw and perfect portion . &                    &                      &                    \\ \cmidrule{2-5} 
                                           & the wine was very average and the food was even less .                      & \multirow{2}{*}{1} & \multirow{2}{*}{5}   & \multirow{2}{*}{5} \\
                                           & the wine was very average and the food was even less .                          &                    &                      &                    \\ \bottomrule
\end{tabular}
\end{table}

\paragraph{Instruction of \textsc{Captions}} 
You are required to evaluate the quality of style transferred sentences in terms of transfer strength, content preservation and language quality.
\begin{description}[style=unboxed,leftmargin=0.2cm]
\item[- \textnormal{Transfer strength:}] You need to judge if the objective style is added to the original sentence and score the extent from 1 to 5 point: If the objective style is perfectly added, score 5 point; If the non-objective style is added, score 1 point; If no style is added, score 3 point.
\item[- \textnormal{Content preservation:}] You need to judge if the content of generated sentence is maintained and score the extent from 1 to 5 point. Higher score suggests better content preservation.
\item[- \textnormal{Language quality:}] You need to judge the quality of generated sentence in terms of grammar, fluency and so on. Higher score suggests better language quality.
\end{description}
Examples in Table \ref{tab:human_eval_example_style} should help you to understand the standard roughly.

\begin{table}[]
\centering
\scriptsize
\caption{Human evaluation examples on style transfer (\textsc{Captions}).}
\label{tab:human_eval_example_style}
\begin{tabular}{llccc}
\toprule
Transfer Direction & Source (top) and transferred (bottom) sentences & \makecell[c]{Transfer\\Strength}  & \makecell[c]{Content\\Preservation} & \makecell[c]{Language\\Quality}   \\ \midrule
\multirow{4}{*}{\makecell[l]{From factual\\to romantic}} & artists working on a mosaic on the ground .                      & \multirow{2}{*}{5} & \multirow{2}{*}{1}   & \multirow{2}{*}{3} \\
                                           & teenagers working on a lawn on the ground by their honeymoon .                     &                    &                      &                    \\ \cmidrule{2-5} 
                                           & brown dog chasing black dog through snow .  & \multirow{2}{*}{4} & \multirow{2}{*}{4}   & \multirow{2}{*}{2} \\
                                           & brown dog playfully black dog through snow .  &                    &                      &                    \\ \midrule
\multirow{4}{*}{\makecell[l]{From factual\\to humorous}} & brown dog chasing black dog through snow .    & \multirow{2}{*}{5} & \multirow{2}{*}{3}   & \multirow{2}{*}{3} \\
                                           & brown dog chasing long stick in the snow , searching for bones . &                    &                      &                    \\ \cmidrule{2-5} 
                                           & some people at a part gather to take a picture .                      & \multirow{2}{*}{3} & \multirow{2}{*}{5}   & \multirow{2}{*}{2} \\
                                           & many people pose for a picture at a birthday show to show .                          &                    &                      &                    \\ \bottomrule
\end{tabular}
\end{table}

\section{Examples}
Table \ref{tab:yelp_examples}, \ref{tab:amazon_examples}, and \ref{tab:cations_examples} present some examples on \textsc{Yelp}, \textsc{Amazon}, and \textsc{Captions}, respectively. Different colors denote the style
of corresponding expressions. Underlined pieces are redundant content expressions that the source
sentence doesn't indicate. We can see that our methods successfully transfer the style and maintain the
content.

\begin{table}[]
\centering
\caption{From \textcolor{red}{negative} to \textcolor{green}{positive} (\textsc{Yelp}).}
\label{tab:yelp_examples}
\begin{tabular}{ll}
\toprule
Source & the wine was very \textcolor{red}{average} and the food was even \textcolor{red}{less} . \\
Reference & the wine was \textcolor{green}{above average} and the food was even \textcolor{green}{better} \\ \midrule
RS & the \uline{food} was \textcolor{green}{good} , but the food was \textcolor{green}{good} . \\
\textsc{CrossAlign} & the wine was very \textcolor{red}{average} and the food was even  \textcolor{red}{less} . \\ 
\textsc{MultiDecoder} & the wine was very \textcolor{red}{average} and the \uline{service} was even \textcolor{green}{better} . \\
Ours (Embedding) & the wine was very \textcolor{green}{good} and the food was even \textcolor{green}{good} . \\ \midrule
\textsc{Frequency} & the wine was pretty \textcolor{green}{good} and the food was even \textcolor{red}{less} . \\
\textsc{Attention} & the wine , very \textcolor{red}{average} and the food is even \textcolor{green}{better} . \\
\textsc{Gradient} & the wine was \textcolor{green}{great} and the food is even \textcolor{green}{better} . \\
\textsc{IntegratGrad} & the wine \uline{selection} and the food is \textcolor{green}{good} . \\
LIME & the \uline{service} very \textcolor{green}{friendly} and the food is \textcolor{green}{good} . \\
Ours (Prototype form) & the wine was very \textcolor{green}{good} and the food was even  \textcolor{green}{better} . \\ \bottomrule

\end{tabular}
\end{table}

\begin{table}[]
\centering
\caption{From \textcolor{red}{negative} to \textcolor{green}{positive} (\textsc{Amazon}).}
\label{tab:amazon_examples}
\begin{tabular}{ll}
\toprule
Source & this is the \textcolor{red}{worst} game i have come across in a long time . \\
Reference & this is the \textcolor{green}{best} game i have come across in a long time \\ \midrule
RS & this \uline{headset} is the \textcolor{red}{worst} i ve had in the past . \\
\textsc{CrossAlign} & this is the \textcolor{green}{best} \uline{sounding bumper} i have in in the past . \\ 
\textsc{MultiDecoder} & this is the \textcolor{green}{best} \uline{spinner that we ve put with this set} . \\
Ours (Embedding) & this is the \textcolor{green}{best} game i ve played in a long time. \\ \midrule
\textsc{Frequency} & i \textcolor{green}{love} that they come across in a long time . \\
\textsc{Attention} & this is the \textcolor{green}{best} \uline{case} i have come across in a long time . \\
\textsc{Gradient} & this is the \textcolor{red}{worst} game i have come in a \uline{kitchen} . \\
\textsc{IntegratGrad} & it is the \textcolor{green}{best} i have come in in a long time . \\
LIME & i \textcolor{green}{like} the fact that i have come across it in a long time . \\
Ours (Prototype form) & this is the \textcolor{green}{best} game i have come across in a long time . \\ \bottomrule

\end{tabular}
\end{table}

\begin{table}[]
\centering
\caption{From factual to \textcolor{blue}{romantic} (\textsc{Captions}).}
\label{tab:cations_examples}
\begin{tabular}{ll}
\toprule
Source & a brown dog and a grey dog play in the grass . \\
Reference & a brown dog and a grey dog in \textcolor{blue}{love} play in the grass . \\ \midrule
RS & \makecell[l]{\uline{a black dog jumping over a pole} to \textcolor{blue}{experience the highs of} \\ \textcolor{blue}{nature} .} \\
\textsc{CrossAlign} & a brown dog and a \uline{brown dog} play in the grass . \\ 
\textsc{MultiDecoder} & a brown dog and a grey dog play in the grass . \\
Ours (Embedding) & \makecell[l]{a brown dog and a grey dog play in the grass , \textcolor{blue}{enjoying the} \\ \textcolor{blue}{happiness of childhood} .} \\ \midrule
\textsc{Frequency} & a brown dog and a grey dog play in the grass . \\
\textsc{Attention} & a brown dog and a grey dog play in the grass . \\
\textsc{Gradient} & a brown dog and a grey dog play \textcolor{blue}{happily in the grass} . \\
\textsc{IntegratGrad} & a brown dog and a grey dog play in the grass . \\
LIME & a brown dog and a grey dog play in the grass . \\
Ours (Prototype form) & a brown dog and a grey dog play in the grass \textcolor{blue}{with full joy} . \\ \bottomrule

\end{tabular}
\end{table}

\end{document}